%% file: main.tex
\def\EndlessExamPreprint{1}
\documentclass{article}
\usepackage{iclr2027_conference,times}
\usepackage{iftex}
\ifxetex
\usepackage{fontspec}
\ifdefined\EndlessExamPreprint
\else
\fi
\fi
\input{math_commands.tex}

\usepackage[hidelinks]{hyperref}
\usepackage{url}
\usepackage{booktabs,tabularx,array,longtable}
\newcolumntype{Y}{>{\raggedright\arraybackslash}X}
\newcolumntype{P}[1]{>{\raggedright\arraybackslash}p{#1}}
\usepackage{graphicx}
\usepackage{placeins}
\usepackage{amsmath,amssymb}
\usepackage{xcolor}
\usepackage{colortbl}
\input{visual_theme}
\usepackage[most]{tcolorbox}
\usepackage{capt-of}
\input{appendix_style}

\input{tables/result_numbers.tex}

\input{tables/tool_numbers.tex}

\title{The Endless Exam: Mathematical Constructions\\from Today's Models toward Superintelligence}

\ifdefined\EndlessExamPreprint
\author{Muhan Zhang\\The University of Texas at Arlington}
\iclrfinalcopy
\else
\author{Anonymous authors\\Paper under double-blind review}
\fi

\begin{document}
\maketitle
\ifdefined\EndlessExamPreprint
\lhead{}\renewcommand{\headrulewidth}{0pt}
\fi

\begin{abstract}
We introduce the Endless Exam, a benchmark spanning fourteen parameterised families of mathematical construction problems, with verifiable scores that distinguish progress before and beyond published mathematical frontiers. Each submitted object is checked automatically for validity and assigned a relative quality score against a published frontier or construction baseline, without capping improvements at $1$. The benchmark draws long-term challenges from open mathematical problems and generates larger instances by varying their parameters. Compact certificates allow large constructions to be verified without listing every element. Across nine models evaluated on \DistinctN{} distinct instances, continuous quality scores distinguish performance even though none of the 30 published-frontier references is surpassed. Size--quality curves show how construction quality changes as problem size increases. We release the generators, verifiers, references, model responses and analysis to support continued measurement before and beyond human frontiers.
\end{abstract}

\section{Introduction}

Measuring progress toward artificial superintelligence (ASI) requires both scores that distinguish advances beyond the best human answers and tasks that continue to offer room for improvement. Mathematical construction problems meet these needs because finding a better object can demand a new idea, while its validity can be checked and its quality measured automatically. Varying the problem's parameters then provides a continuing supply of harder instances.

Benchmark saturation makes this need immediate: a recent study finds that 29 of 60 language-model benchmarks exhibit high or very high saturation, limiting their ability to distinguish leading models~\citep{akhtar2026saturation}. ARC Prize recently reported $99.9\%$ for GPT-6 Astra on ARC-AGI-3 Semi-Private with its Provider Adapter harness~\citep{kamradt2026astra}.\footnote{The $99.9\%$ result uses high reasoning effort. With the Standard evaluation setup, the same report gives $62.7\%$ at max effort.} Once models reach benchmark ceilings, evaluation must continue to distinguish their progress and offer harder instances.

\begin{figure}[t]
\centering
\includegraphics[width=\linewidth]{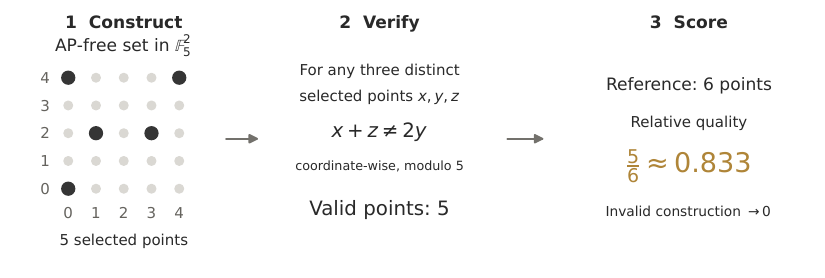}
\caption{An illustrative AP-free set in $\mathbb F_5^2$. Here $x,y,z$ denote distinct selected points, with coordinate-wise arithmetic modulo 5. Dividing the five valid points by the six-point optimum gives a relative quality of $5/6\approx0.833$.}
\label{fig:framework}
\end{figure}

We introduce the \emph{Endless Exam}, a benchmark built from fourteen parameterised families of mathematical construction problems, with 69 evaluation instances. Consider a progression-free set: a collection of points in $\mathbb{F}_5^n$ containing no three distinct points $x,y,z$ with $x+z=2y$ (Figure~\ref{fig:framework}). Given a proposed set, the evaluator checks for forbidden triples and counts its points, rewarding larger valid sets even when they exceed every published construction. Changing the dimension produces a new task. We apply the same approach to graphs and codes, checking each construction and scoring its quality.

In many of these families, the best published constructions still leave a substantial gap to the proven upper bounds. For seven-colour Schur partitions and progression-free sets in $\mathbb F_5^6$, these bounds are $\SchurBoundRatio$ and $\APBoundRatio$ times the best published construction sizes, respectively, and the optimal sizes remain unknown.

To assess current model performance on these construction problems, we evaluate nine models in fourteen tool-free configurations: GPT-6 Astra, GPT-5.6 Luna, Claude Fable 5.1, Claude Opus 5.5, DeepSeek V4.1 Flash, Qwen3.8-27B and Qwen3.5 at 4B, 9B and 27B. To examine the effect of tool access, we additionally evaluate Astra, Luna and Opus 5.5 at high effort with code and web access on the 69 instances.

Our contributions are:
\begin{enumerate}
\item \textbf{An uncapped measure of mathematical progress.} Relative quality puts diverse construction objectives on a common scale and preserves the magnitude of improvements beyond a reference.
\item \textbf{An expandable suite of construction problems.} Fourteen families generate new instances by varying parameters such as dimension, graph degree and code length; we evaluate \DistinctN{} instances.
\item \textbf{Automatic verification at large scales.} We verify constructions through their mathematical properties, even when the optimum is unknown. Compact descriptions and algebraic certificates support evaluation of objects too large to enumerate, including submitted codes with over a trillion codewords (Section~\ref{sec:formats}).
\end{enumerate}

The Endless Exam is an open-source benchmark, and we welcome community contributions of new constructions, task families and verifier improvements.
\ifdefined\EndlessExamPreprint
Code and data are available at \mbox{\url{https://github.com/ricolalemon/endless-exam}}.
\fi

\section{Related work}
\label{sec:related}

\paragraph{Mathematical reasoning benchmarks.}
Humanity's Last Exam~\citep{phan2025hle} and FrontierMath~\citep{glazer2024frontiermath,epoch2026frontiermath} assess expert-level problem solving. IMProofBench~\citep{schmitt2025improofbench} and First Proof~\citep{abouzaid2026firstproof} evaluate research-level proofs, IMO-AnswerBench~\citep{luong2025imobench} evaluates Olympiad answers, and ARC-AGI~\citep{chollet2019arc,chollet2025arcagi2} tests abstract reasoning. Table~\ref{tab:benchmark_comparison} compares representative mathematical benchmarks and problem collections.

\paragraph{Verifiable constructions and problem generation.}
MathConstruct~\citep{balunovic2025mathconstruct} checks constructed objects and varies problem parameters to assess robustness, while MathConstraint~\citep{pati2026mathconstraint} generates constraint problems, uses solvers to determine whether solutions exist, and checks submitted constructions. The Endless Exam assigns graded scores to valid constructions: for the same instance, more set elements, more graph vertices or fewer scalar multiplications receive higher scores. This lets us measure how much constructions improve at a given size and how their quality changes at larger sizes.

\paragraph{Open mathematical problems.}
FrontierMath: Open Problems and the Erd\H{o}s Problems and Optimization Constants collections track unsolved questions and mathematical bounds~\citep{epoch2026openproblems,bloom2021erdosproblems,davis2026constants}. HorizonMath~\citep{horizonmath2026} evaluates open problems with automated verification and reports the proportion solved. EternalMath~\citep{ma2026eternalmath} derives executable, parameterised tasks from mathematical literature, while Formal Conjectures~\citep{firsching2026formal} collects research problems formalised in Lean for proof discovery and autoformalisation. The Endless Exam scores construction quality on an \emph{uncapped scale}, distinguishing the magnitude of improvements even after a published frontier is surpassed.

\begin{table}[t]
\centering\footnotesize\setlength{\tabcolsep}{2.5pt}
\renewcommand{\arraystretch}{1.06}
\input{tables/benchmark_comparison}
\caption{Comparison of mathematical benchmarks and problem collections, using the cited releases. Unsolved indicates inclusion of open research targets. Our 69 instances span 14 families. Scalable denotes larger instances generated by increasing parameters under the same rules. $^{*}$IMProofBench's main proofs are human-graded; follow-up answers are automatically checked.}
\label{tab:benchmark_comparison}
\end{table}

\paragraph{Mathematical discovery systems.}
FunSearch~\citep{romera2024funsearch}, AlphaEvolve~\citep{novikov2025alphaevolve} and AlphaTensor~\citep{fawzi2022alphatensor} have improved mathematical constructions. AlphaEvolve evolves programs using numerical quality feedback, and its mathematical study spans 67 problems with generalisation across parameters~\citep{georgiev2025exploration}. To compare such progress across systems, families and problem sizes, the Endless Exam combines relative quality, an uncapped overall score and size--quality curves within a common evaluation framework.

\section{The exam}
\label{sec:exam}

\subsection{Design principles}
\label{sec:principles}
We select construction families that support continued measurement through improving construction quality and increasing problem size. Each task specifies the conditions a construction must satisfy and an objective to optimise, such as maximising the number of graph vertices or minimising the number of scalar multiplications in a matrix multiplication algorithm. Among valid constructions, every improvement in the objective raises the score, including improvements beyond the reference. The evaluator checks each construction's mathematical properties and computes its objective value directly, without a model judge. These checks do not require knowing the optimum, allowing progress to be measured even on open problems. Basing scores on the submitted object also makes them resistant to reward hacking through persuasive prose or unsupported quality claims. Compact descriptions and certificates extend this approach to objects too large to enumerate. Varying the parameters creates new instances under the same mathematical rules; larger instances admit more possible constructions and can make high-quality ones harder to find, providing more demanding tasks as models advance.

\subsection{Families and instances}
\label{sec:tiers}
The fourteen families span forbidden-pattern sets, geometry, graphs, codes and algebraic designs (Table~\ref{tab:families}). We evaluate the \DistinctN{} instances of benchmark release \texttt{bench-v1.0}. To our knowledge, the mathematical optimum is unknown for each of the \DistinctN{} main-evaluation instances. For \PanelOneN{} instances, \emph{published frontiers} provide a reference for measuring how closely models approach existing mathematical results and whether they improve on them. For the remaining \PanelTwoN{}, we use verified constructions generated by the algorithms described in Appendix~\ref{app:lit} as reproducible baselines. These baselines may be weaker than published constructions. Every reference is supported by a verified construction and remains unchanged within \texttt{bench-v1.0}. Both groups use the same relative-quality score, which increases as constructions improve, including beyond the reference (Section~\ref{sec:scoring}).

\input{representative_tasks}

\subsection{Answer formats and verification}
\label{sec:formats}
The examples in Section~\ref{sec:representative_tasks} illustrate two accepted forms of answer: an explicit object and a compact description. Across the suite, compact forms include products, lattice orbits, digit sets, difference sets, Cayley generators and algebraic certificates. The verifier either expands the description and checks the resulting object, or verifies the components and algebraic conditions that establish its validity and size. In our evaluations, checking the finite factors of a submitted Shannon code certifies over a trillion codewords (Appendix~\ref{case:shannon}).

The primary verifier checks whether each submitted construction satisfies the mathematical constraints and computes its objective value. For every construction it judges valid, a second verifier checks the same constraints and recomputes the objective value using a separate implementation of the mathematical checks. The two verifiers agree on validity and objective value for all these model constructions and all 69 reference constructions. We also test the verifiers on deliberately invalid constructions and compare their decisions with exhaustive checks on small Shannon and trifference instances. They reject all invalid test cases and agree with the exhaustive checks (Appendix~\ref{app:verifier_validation}). Appendix~\ref{app:formats} gives the certificate checks and their size and time limits.

\subsection{Evaluation protocol}
\label{sec:protocol}
All configurations receive the same instances, with prompts specifying the problem, its parameters and the required answer format (Appendix~\ref{app:tiers}). In the tool-free track, models produce one response per instance within a budget of 128k output tokens, including reasoning tokens. Invalid responses and generation-budget exhaustion score zero. Each effort level is evaluated as a separate configuration (Appendix~\ref{app:runner}).

The tool-assisted track gives GPT-6 Astra, GPT-5.6 Luna and Claude Opus 5.5, all at high effort, code execution and web access while retaining the mathematical prompts and answer formats. Each evaluation has four CPU threads, 16\,GiB of memory and a two-hour wall-clock limit, with no limit on total output tokens. When an evaluation finishes or reaches a resource limit, we assess the construction saved before the deadline. Valid constructions receive their quality scores; missing or invalid submissions receive zero (Appendix~\ref{app:tools}).

\section{Scoring}
\label{sec:scoring}

\subsection{Relative quality and overall score}
We measure \emph{relative quality} with the same formula for both reference groups. For parameters $p$, verified objective $a$ and reference $h(p)$,
\[
r(a,p)=
\begin{cases}
a/h(p), & \text{valid maximisation answer},\\
h(p)/a, & \text{valid minimisation answer},\\
0, & \text{invalid or empty response}.
\end{cases}
\]
Relative quality $1$ matches the reference, and every further improvement increases the ratio without clipping it at $1$. References remain unchanged within each benchmark version, so successive improvements remain comparable.

For instances with a published frontier, $h(p)$ is the objective value of the cited construction. For the remaining instances, $h(p)$ is the best objective value among the verified constructions produced by the algorithms in Appendix~\ref{app:lit}. Relative quality above $1$ indicates improvement over the reference; surpassing a published frontier yields a candidate mathematical record. Appendix~\ref{app:lit} documents the references and baseline procedures.

Published eight-colour Schur constructions illustrate how scores distinguish successive improvements. Against the historical reference of 5,041 elements derived from \citet{fredricksen2000schur}, constructions of sizes 5,286~\citep{rowley2021templates} and 5,362~\citep{bengone2026schur} receive scores of $1.049$ and $1.064$, respectively. Both surpass the reference, but the scores distinguish their quality. The current benchmark uses 5,362 as its eight-colour reference (Appendix~\ref{case:schur}).

The \textbf{overall score} is $S=(100/69)\sum_{i=1}^{69}r_i$, where $r_i$ is the relative quality for instance $i$. Invalid submissions receive zero, and a score of 100 matches the references on average. Table~\ref{tab:headline} also reports mean relative quality for each reference group, with $95\%$ confidence intervals obtained by bootstrapping instances within that group.

Each construction instance provides a richer picture of performance than a binary correct/incorrect label. Verification establishes whether the answer is valid, and relative quality measures how well the construction performs against its published frontier or construction baseline. For example, valid constructions with relative quality scores of $0.5$, $0.9$ and $1.2$ would all pass a validity check, but receive different quality scores. Our \DistinctN{} instances therefore reveal more than \DistinctN{} success-or-failure outcomes: they show how close each construction comes to its reference and how far it improves beyond it.

\subsection{Gap closed}
\label{sec:headroom}
Gap closed measures progress from the reference $h$ to a proven bound $b$. For maximisation it is $\max(0,\ln(a/h)/\ln(b/h))$. For minimisation, the ratios $a/h$ and $b/h$ are replaced by $h/a$ and $h/b$. Constructions that do not improve on the reference score $0$, while reaching the bound scores $1$; invalid answers also score $0$. We average gap closed over the \GapN{} instances with proven bounds: \GapPublishedN{} use published frontiers and \GapConstructionN{} use construction baselines. The \TargetN{} LABS instances use a conjectured target and are reported separately (Appendix~\ref{app:gap_breakdown}).

\begin{figure}[t]
\centering
\includegraphics[width=\linewidth]{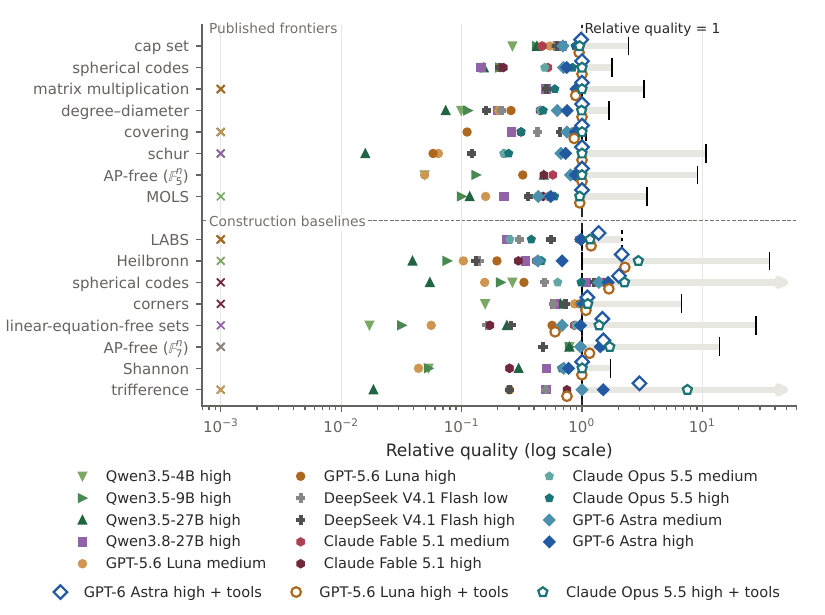}
\caption{Mean relative quality across fourteen families. Model points and bar endpoints are normalised to the reference and averaged over the same instances in each row; model means include zero scores. Upper rows use published frontiers; lower rows use construction baselines. A ratio of $1$ matches the reference. Grey bars extend from the reference level of $1$ to mathematical bounds or conjectured targets. A solid mark at the right end indicates a proven bound; a dashed mark indicates a conjectured target. An arrow means the endpoint lies beyond the plotted range. Filled markers indicate tool-free evaluations and hollow markers tool-assisted evaluations. Crosses mark means below 0.001. AP-free: arithmetic-progression-free; LABS: low-autocorrelation binary sequences; MOLS: mutually orthogonal Latin squares.}
\label{fig:scale}
\end{figure}

\section{Results}
\label{sec:results}

\begin{table}[t]
\centering\footnotesize\setlength{\tabcolsep}{3pt}
\input{tables/headline_table}
\caption{Results on 69 instances. The final three rows use code and web access (Section~\ref{sec:protocol}). Overall score is uncapped. Brackets show $95\%$ confidence intervals. Validity uses all 69 responses; gap closed uses \GapN{} instances with proven bounds (Section~\ref{sec:headroom}).}
\label{tab:headline}
\end{table}

\begin{figure}[t]
\centering
\includegraphics[width=\linewidth]{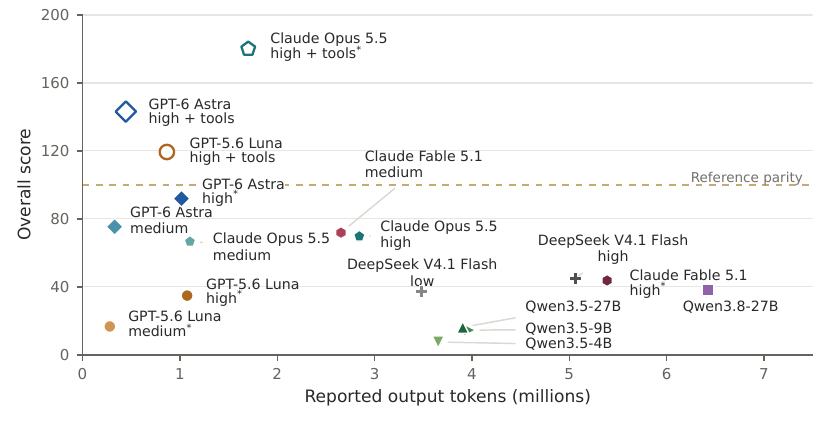}
\caption{Overall score and output tokens for 69 instances, including reasoning and all model turns. For models marked with an asterisk, some output-token counts are unavailable, so the plotted totals are lower bounds. All Qwen models use high effort; tokenisation varies by model.}
\label{fig:score_tokens}
\end{figure}

\subsection{Distinguishing today's models}
Tool-free overall scores span $\ScoreMin$--$\ScoreMax$ (Table~\ref{tab:headline}). On the 30 instances with published frontiers, mean relative quality spans $\POneMin$--$\POneMax$ despite a $\BinaryBreakthroughMaxPct\%$ breakthrough rate for all fourteen configurations. GPT-6 Astra at high effort reaches $\AstraHighPOne$ ($95\%$ CI $\AstraHighCILow$--$\AstraHighCIHigh$). Mean relative quality on the 39 instances with construction baselines ranges from $\PTwoMin$ to $\PTwoMax$. Astra high also achieves the largest tool-free mean gap closed, at $\HeadroomMaxPct\%$ across the \GapN{} instances with proven bounds. Figure~\ref{fig:scale} and Appendix~\ref{app:more} show the differences across families.

Both model size and reasoning effort affect quality. On instances with published frontiers, Qwen3.5's mean relative quality rises from $\QwenFourPOne$ to $\QwenNinePOne$ to $\QwenTwentySevenPOne$ at 4B, 9B and 27B (Figure~\ref{fig:model_size}). Raising effort from medium to high increases Astra's mean from $\AstraMedPOne$ to $\AstraHighPOne$ and GPT-5.6 Luna's from $\LunaMedPOne$ to $\LunaHighPOne$. Claude Fable 5.1's overall score instead falls from $\FableScore$ at medium effort to $\FableHighScore$ at high effort, where \FableHighBudgetStops{} of the \DistinctN{} responses exhaust the 128k output budget. Additional reasoning effort can therefore consume the budget without producing a completed construction. Opus 5.5 scores $\OpusMediumScore$ at medium effort and $\OpusHighScore$ at high effort, with \OpusMediumValid{} and \OpusHighValid{} valid answers, respectively. Figure~\ref{fig:score_tokens} relates overall scores to output tokens: Astra medium outscores Fable 5.1 medium with fewer tokens, while Qwen3.8-27B uses the most reported output tokens but scores below both.
\par

\paragraph{Code and web access.}
GPT-6 Astra at high effort produces valid constructions on all \DistinctN{} instances without tools. With code and web access, it improves their average quality, raising its overall score from $\AstraHighScore$ to $\ToolScore$ while maintaining $100\%$ validity. GPT-5.6 Luna's score at high effort rises from $\LunaToolFreeScore$ to $\LunaToolScore$ as its valid answers increase from \LunaToolFreeValid{} to \LunaToolValid{}. On the \LunaBothValidN{} instances with valid answers in both tracks, its mean relative quality also rises from $\LunaBothValidFreeQuality$ to $\LunaBothValidToolQuality$. Luna therefore improves both its ability to produce valid answers and the quality of those answers. Claude Opus 5.5 rises from $\OpusHighScore$ to $\OpusToolScore$, with validity increasing from \OpusHighValid{} to \OpusToolValid{} instances.

On the \PanelOneN{} instances with published frontiers, tool-assisted Astra, Luna and Opus 5.5 match \ToolPublishedMatched{}, \LunaToolPublishedMatched{} and \OpusToolPublishedMatched{} references, respectively, without surpassing any. The evaluated models already match many published frontiers in this group, but have yet to surpass one. On the \PanelTwoN{} instances with construction baselines, Astra, Luna and Opus 5.5 improve on the construction baselines for \ToolConstructionAbove{}, \LunaToolConstructionAbove{} and \OpusToolConstructionAbove{} instances (Appendix~\ref{app:tools}). Opus 5.5 achieves the highest overall score, with its advantage concentrated in Heilbronn and the length-64 trifference instance, while Astra achieves the highest mean relative quality on published-frontier instances. These improvements already demonstrate scoring beyond $1$; future improvements over published frontiers would be measured in the same way.

\subsection{Quality as problem size increases}
\label{sec:ladder}
With tools at high effort, mean gap closed remains small: $\ToolGapClosedPct\%$ for GPT-6 Astra, $\LunaToolGapClosedPct\%$ for GPT-5.6 Luna and $\OpusToolGapClosedPct\%$ for Claude Opus 5.5 across the \GapN{} instances with proven bounds. Despite improvements over some references, the models close only a small fraction of the logarithmic gap to the proven bounds, leaving substantial room for further improvement under this metric.

If future models close a substantial fraction of these gaps and the existing instances become easy, larger parameter settings can provide more demanding tasks under the same mathematical rules. The name \emph{Endless Exam} reflects these two routes for continued measurement: improving construction quality within an instance and extending the task to larger dimensions, graph diameters or codeword lengths as models advance.

\begin{figure}[t]
\centering
\includegraphics[width=\linewidth]{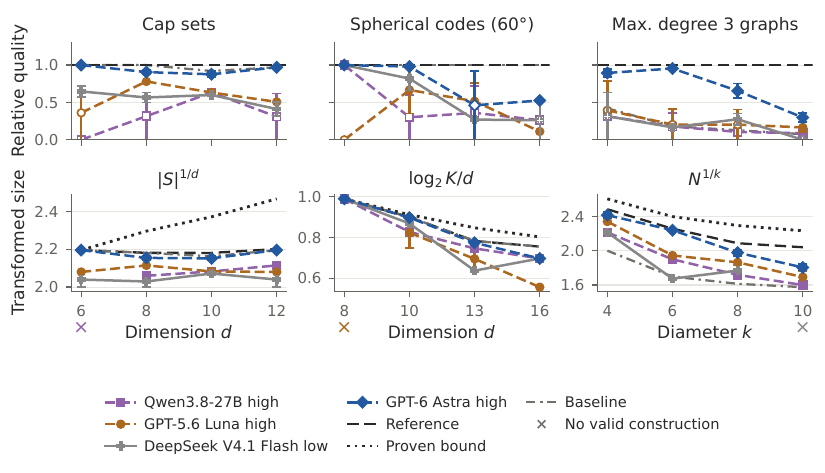}
\caption{Size--quality curves with 128k output tokens. Upper panels show mean relative quality over two independent responses per size, including zeroes; whiskers show their range and hollow markers indicate an invalid response. Grey dash-dot lines show baseline/reference ratios. Lower panels rescale valid construction sizes; whiskers span valid responses and crosses indicate none. Here $d$ is dimension, $k$ graph diameter, $|S|$ cap-set size, $K$ spherical-code size and $N$ graph order; $|S|^{1/d}$ gives the exponential growth base. Appendix~\ref{app:ladder_results} gives the full comparison.}
\label{fig:ladder}
\end{figure}

To examine how current construction methods scale, we evaluate four configurations: DeepSeek V4.1 Flash at low effort, and Qwen3.8-27B, GPT-6 Astra and GPT-5.6 Luna at high effort. Each configuration is tested at four sizes in five benchmark families and an additional integer progression-free task, with two independent responses per size and the same token budget. Size--quality curves show whether construction quality keeps pace with the reference as the task grows. Figure~\ref{fig:ladder} shows three families; Appendix~\ref{app:more} gives all six tasks and an additional graph series with maximum degree four.

Larger graph and spherical-code instances expose widening gaps to the reference constructions. For graphs of maximum degree three, all four models have lower mean relative quality at diameter 10 than at diameter 4. At diameter 10, the means are $\CubicLFourAstra$ for Astra, $\CubicLFourLuna$ for Luna, $\CubicLFourQwen$ for Qwen3.8-27B and $\CubicLFourDS$ for DeepSeek V4.1 Flash. Astra, Luna and Qwen3.8-27B still return valid graphs at this size, but their constructions fall well below the reference. All four models' spherical-code scores also fall between dimensions 10 and 16. On the larger instances, some models fail to produce valid constructions, while others return valid constructions that fall further below the reference.

The gap to a proven bound can also widen while a model stays close to the reference. For cap sets in six dimensions, \CapLOneSize{} points meet both the reference and the proven upper bound. In twelve dimensions, both Astra answers combine two such caps to obtain \CapLFourSize{} points, reaching relative quality $\CapLFour$ against the reference of \CapLFourReference{} points. The proven upper bound is \CapLFourBound{}, so these answers reach only $\CapLFourBoundPct\%$ of that bound. The gap between the reference and the proven upper bound also grows across the graph and Schur size series (Appendix~\ref{app:more}). These bounds need not be attainable; their separation from known constructions marks an unresolved mathematical gap at larger sizes.

\subsection{Sensitivity to computational search budget}
\label{sec:search}
The reference searches also let us examine how much construction quality improves with additional computation. For these procedures, eight families are classified as \emph{search-resistant}, with strong constructions based on products, lattices, digit sets or recursions, while five may benefit from direct search. Across all eight search-resistant families, increasing the search budget from 10 to 600 seconds improves the baseline by less than $10\%$ on every measured instance, while Heilbronn benefits more (Figure~\ref{fig:search}; Appendix~\ref{app:search_audit}). Trifference is excluded because its baseline procedure selects and combines existing codes without using the search-time budget.
\par

These measurements describe the tested search algorithms. Tool-assisted models can use broader methods, and discovery systems such as AlphaEvolve can develop new algorithms~\citep{georgiev2025exploration}. As stronger methods narrow the gap to mathematical bounds, future benchmark versions can add larger instances. Each released version retains its original instances and references so that scores remain comparable.

\FloatBarrier
\section{Limitations}
\label{sec:limitations}
Within the available evaluation budget, we test selected models and reasoning-effort levels, using one evaluation per instance and configuration in the main comparison. Repeated evaluations would quantify run-to-run variability, which the instance-bootstrap intervals do not measure. Answer formats and verification budgets also limit the constructions that can be evaluated (Appendix~\ref{app:representation_limits}); future benchmark versions can extend certificate formats alongside instance sizes, while preserving earlier releases for comparison.

\section{Conclusion}
The Endless Exam measures mathematical progress before and beyond published frontiers by combining uncapped scores with parameterised families and compact certificates. At high effort, tools raise the scores of GPT-6 Astra, GPT-5.6 Luna and Claude Opus 5.5 to $\ToolScore$, $\LunaToolScore$ and $\OpusToolScore$, with gains beyond construction baselines while none of the 30 published-frontier references is surpassed. Historical Schur constructions show how scores distinguish subsequent breakthroughs, and size--quality curves extend the comparison to larger tasks, providing ways to continue measurement toward superintelligence.
\par
\label{page:mainend}

\clearpage
\ifdefined\EndlessExamPreprint
\section*{Acknowledgments}
The author thanks Prof. Peiran Yu for providing computational resources for the experiments.
\fi

\section*{Reproducibility statement}
\ifdefined\EndlessExamPreprint
Our repository provides generators, verifiers, reference constructions, model responses, inference settings and analysis scripts to reproduce every score, table and figure.
\else
The supplementary material provides anonymised generators, verifiers, reference constructions, model responses, inference settings and analysis scripts to reproduce every score, table and figure.
\fi Instances are specified by family, parameters and random seed. Published constructions link to their sources; independent verifiers and regression tests accompany the release.
\ifdefined\EndlessExamPreprint
Large language models assisted with literature discovery, code development, mathematical derivations and verification, analysis and writing; the author is responsible for the final content.
\else
\section*{AI use statement}
The author led the study and is responsible for the final content. AI tools assisted with literature search, research design, implementation, mathematical derivations, analysis and writing.
\fi

\bibliography{references}
\bibliographystyle{iclr2027_conference}

\clearpage
\appendix
\input{appendix/appendix}

\end{document}

%% file: math_commands.tex
\usepackage{amsmath,amsfonts,bm}

\def\eqref#1{equation~\ref{#1}}
\def\1{\bm{1}}

\DeclareMathAlphabet{\mathsfit}{\encodingdefault}{\sfdefault}{m}{sl}
\SetMathAlphabet{\mathsfit}{bold}{\encodingdefault}{\sfdefault}{bx}{n}

%% file: visual_theme.tex
\definecolor{paperink}{HTML}{252525}
\definecolor{paperaccent}{HTML}{AF8538}
\definecolor{papermuted}{HTML}{73716C}
\definecolor{paperline}{HTML}{DAD8D3}
\definecolor{paperwhite}{HTML}{FFFFFF}
\definecolor{paperwarmgray}{HTML}{F1F0ED}
\definecolor{paperresult}{HTML}{FFFCF6}
\definecolor{paperresulttitle}{HTML}{F4ECD9}

%% file: appendix_style.tex
\colorlet{appendixink}{paperink}
\colorlet{appendixgold}{paperaccent}
\tcbset{appendixbase/.style={enhanced,boxrule=.35pt,arc=1.4pt,
  left=8pt,right=8pt,top=6pt,bottom=6pt,before skip=8pt,after skip=8pt,
  fonttitle=\bfseries,coltitle=black}}
\newtcolorbox{taskbox}[1][]{appendixbase,colback=paperwhite,colframe=black!22,
  colbacktitle=paperwarmgray,#1}
\newtcolorbox{constructionbox}[1][]{appendixbase,colback=paperwhite,
  colframe=appendixink!35,colbacktitle=paperwarmgray,#1}
\newtcolorbox{resultbox}[1][]{appendixbase,colback=paperresult,
  colframe=appendixgold!45,colbacktitle=paperresulttitle,#1}
\newcommand{\catalogueheading}[2]{\refstepcounter{subsection}\label{family:#1}%
  \addcontentsline{toc}{subsection}{\protect\numberline{\thesubsection}#2}%
  \noindent{\large\bfseries\color{appendixink}\thesubsection\quad #2}\par\smallskip}
\newcommand{\appendixpage}{\par\clearpage}

\newcommand{\readingnote}[1]{\begin{constructionbox}#1\end{constructionbox}}

%% file: tables/result_numbers.tex
\newcommand{\PanelOneN}{30}
\newcommand{\PanelTwoN}{39}
\newcommand{\DistinctN}{69}
\newcommand{\POneMin}{0.07}
\newcommand{\POneMax}{0.76}
\newcommand{\ScoreMin}{7.55}
\newcommand{\ScoreMax}{91.90}
\newcommand{\AstraHighScore}{91.90}

\newcommand{\FableScore}{71.90}
\newcommand{\FableHighScore}{43.73}
\newcommand{\OpusHighScore}{69.81}
\newcommand{\OpusMediumScore}{66.70}
\newcommand{\OpusHighValid}{57}
\newcommand{\OpusMediumValid}{61}

\newcommand{\FableHighBudgetStops}{32}
\newcommand{\BinaryBreakthroughMaxPct}{0}
\newcommand{\PTwoMin}{0.08}
\newcommand{\PTwoMax}{1.04}
\newcommand{\AstraHighPOne}{0.76}
\newcommand{\AstraMedPOne}{0.66}

\newcommand{\AstraHighCILow}{0.69}
\newcommand{\AstraHighCIHigh}{0.83}

\newcommand{\AstraMedValidCalls}{68}
\newcommand{\AstraHighValidCalls}{69}
\newcommand{\HeadroomMaxPct}{1.6}
\newcommand{\LunaMedPOne}{0.16}
\newcommand{\LunaHighPOne}{0.25}

\newcommand{\TaskGraphVertices}{203}
\newcommand{\TaskGraphReference}{364}
\newcommand{\TaskGraphRatio}{0.558}
\newcommand{\GapN}{64}
\newcommand{\GapPublishedN}{30}
\newcommand{\GapConstructionN}{34}
\newcommand{\TargetN}{5}
\newcommand{\QwenFourPOne}{0.07}
\newcommand{\QwenNinePOne}{0.12}
\newcommand{\QwenTwentySevenPOne}{0.13}
\newcommand{\SchurBoundRatio}{8.1}
\newcommand{\APBoundRatio}{11.6}
\newcommand{\CapLOneSize}{112}
\newcommand{\CapLFourSize}{12,544}
\newcommand{\CapLFourReference}{12,928}
\newcommand{\CapLFourBound}{50,571}
\newcommand{\CapLFourBoundPct}{24.8}

\newcommand{\CapLFour}{0.97}

\newcommand{\QuarticMTwo}{0.53}
\newcommand{\QuarticMFour}{0.15}
\newcommand{\CubicLFourLuna}{0.17}
\newcommand{\CubicLFourQwen}{0.09}
\newcommand{\CubicLFourDS}{0.00}
\newcommand{\CubicLFourAstra}{0.30}

\newcommand{\AstraEffN}{25}
\newcommand{\AstraMedATwoHFR}{0.87}

\newcommand{\AstraHighATwoHFR}{1.01}

\newcommand{\AstraXhighATwoHFR}{1.06}

%% file: tables/tool_numbers.tex
\newcommand{\ToolScore}{143.16}
\newcommand{\ToolPublishedHFR}{0.998}
\newcommand{\ToolConstructionRatio}{1.765}
\newcommand{\ToolPublishedMatched}{29}

\newcommand{\ToolConstructionAbove}{33}
\newcommand{\ToolImproved}{62}
\newcommand{\ToolTied}{7}
\newcommand{\ToolValid}{69}
\newcommand{\ToolGapClosedPct}{6.3}
\newcommand{\ToolTriWords}{177{,}147}
\newcommand{\ToolTriRatio}{9}
\newcommand{\LunaToolScore}{119.35}
\newcommand{\LunaToolFreeScore}{34.85}
\newcommand{\LunaToolPublishedHFR}{0.961}
\newcommand{\LunaToolConstructionRatio}{1.372}
\newcommand{\LunaToolPublishedMatched}{23}
\newcommand{\LunaToolConstructionAbove}{25}
\newcommand{\LunaToolValid}{66}
\newcommand{\LunaToolImproved}{61}
\newcommand{\LunaToolTied}{5}
\newcommand{\LunaToolDeclined}{3}
\newcommand{\LunaToolBeatsAstra}{7}
\newcommand{\LunaToolFreeValid}{47}
\newcommand{\LunaBothValidN}{45}
\newcommand{\LunaBothValidFreeQuality}{0.50}
\newcommand{\LunaBothValidToolQuality}{1.32}
\newcommand{\LunaToolGapClosedPct}{4.7}
\newcommand{\OpusToolScore}{180.06}
\newcommand{\OpusToolPublishedHFR}{0.985}
\newcommand{\OpusToolConstructionRatio}{2.428}
\newcommand{\OpusToolPublishedMatched}{23}
\newcommand{\OpusToolConstructionAbove}{33}
\newcommand{\OpusToolValid}{69}
\newcommand{\OpusToolImproved}{62}
\newcommand{\OpusToolTied}{7}

\newcommand{\OpusToolGapClosedPct}{7.5}
\newcommand{\OpusToolTriWords}{531{,}441}
\newcommand{\OpusToolTriRatio}{27}

%% file: tables/benchmark_comparison.tex
\newcommand{\benchyes}{\textcolor{paperink}{$\checkmark$}}
\newcommand{\benchno}{\textcolor{papermuted}{$\times$}}
\begingroup
\fontsize{8.5}{10}\selectfont
\setlength{\tabcolsep}{3.2pt}
\newcommand{\benchhead}[1]{\begin{tabular}[c]{@{}c@{}}#1\end{tabular}}
\begin{tabularx}{\linewidth}{@{}Yccccccc@{}}
\toprule
Dataset & Problems & Difficulty & Unsolved & Non-proof & \benchhead{Open\\eval.} & \benchhead{Auto-\\check} & \benchhead{Uncapped\\score} \\
\midrule
FrontierMath & 338 & Univ.--Res. & \benchno & \benchyes & \benchno & \benchyes & \benchno \\
FrontierMath: Open Problems & 50 & Research & \benchyes & \benchyes & \benchno & \benchyes & \benchno \\
IMProofBench & 77 & Research & \benchyes & \benchno & \benchno & \benchno$^{*}$ & \benchno \\
First Proof & 10 & Research & \benchno & \benchno & \benchno & \benchno & \benchno \\
Erd\H{o}s Problems & 1,217 & Research & \benchyes & \benchno & \benchno & \benchno & \benchno \\
IMO-AnswerBench & 400 & Olympiad & \benchno & \benchyes & \benchyes & \benchyes & \benchno \\
Opt. Constants & 115 & Research & \benchyes & \benchyes & \benchno & \benchno & \benchno \\
MathConstruct & 127 & Olympiad & \benchno & \benchyes & \benchyes & \benchyes & \benchno \\
MathConstraint & 329 & Tunable & \benchno & \benchyes & \benchyes & \benchyes & \benchno \\
HorizonMath & 113 & Research & \benchyes & \benchyes & \benchyes & \benchyes & \benchno \\
\midrule
\rowcolor{paperresulttitle}\textbf{Endless Exam} & \textbf{69} & Research, scalable & \benchyes & \benchyes & \benchyes & \benchyes & \benchyes \\
\bottomrule
\end{tabularx}
\endgroup

%% file: representative_tasks.tex
\subsection{Representative tasks}
\label{sec:representative_tasks}
These three examples are drawn from the \DistinctN{} evaluation instances and illustrate graphs, geometry and compactly represented codes. Reference values appear here to explain scoring; model prompts contain only the task and answer requirements.

\begingroup
\tcbset{mainexample/.style={arc=0pt,boxrule=.35pt,boxsep=0pt,left=4pt,right=4pt,
 top=2pt,bottom=2pt,before skip=5pt,after skip=5pt,colbacktitle=paperwhite,
 toptitle=2pt,bottomtitle=2pt,titlerule=0pt}}
\begin{taskbox}[mainexample,title={Degree--diameter graphs: degree 4, diameter 5}]
\textbf{Task.} Construct an undirected graph with as many vertices as possible, subject to two constraints: each vertex has at most four neighbours, and any two vertices can be connected by a path of at most five edges. Self-loops and repeated edges are not allowed.

\textbf{Answer and verification.} An edge list specifies the graph. The verifier checks vertex degrees and shortest-path distances, then counts the vertices of a valid graph. GPT-6 Astra at high effort without tools submits a graph with \TaskGraphVertices{} vertices, compared with the published frontier of \TaskGraphReference{}. Its relative quality is therefore $\TaskGraphVertices/\TaskGraphReference\approx\TaskGraphRatio$ (Section~\ref{sec:scoring}).
\end{taskbox}
\begin{taskbox}[mainexample,title={Heilbronn triangles: 46 points in a square}]
\textbf{Task.} Place 46 distinct points in the unit square so that the smallest triangle formed by any three points has the largest possible area. Each coordinate must be a multiple of $10^{-4}$.

\textbf{Answer and verification.} Points are submitted as pairs of integers between 0 and 10,000, scaled by $10^{-4}$. The verifier checks their count, distinctness and range, then computes the minimum triangle area using exact integer arithmetic and the coordinate scale. A larger minimum area scores higher. This instance is evaluated against a construction baseline, illustrating how verified geometric quality provides a continuous score.
\end{taskbox}
\begin{taskbox}[mainexample,title={Trifference codes: ternary strings of length 64}]
\textbf{Task.} Find as many distinct length-64 strings over $\{0,1,2\}$ as possible. For every three distinct strings, there must be at least one position where their symbols are all different: one is 0, one is 1 and one is 2. The objective is the number of strings satisfying this condition together.

\textbf{Answer and verification.} A generator matrix describes all linear combinations of its rows modulo 3. The verifier checks the three-string condition algebraically and counts codewords from the number of independent rows, without enumerating the full code. For this instance, GPT-6 Astra at high effort with tools submits an $11\times64$ matrix with 11 independent rows. Choosing a coefficient of 0, 1 or 2 for each row gives $3^{11}=\ToolTriWords$ distinct strings. This is \ToolTriRatio{} times the construction baseline, so its relative quality is \ToolTriRatio{}. Appendix~\ref{app:trifference_certificates} describes the certificate check.
\end{taskbox}
\endgroup

%% file: tables/headline_table.tex
\begin{tabular}{@{}lrrrrr@{}}
\toprule
& & \multicolumn{2}{c}{Mean relative quality} & & \\
\cmidrule(lr){3-4}
configuration & \shortstack{Overall\\score} & \shortstack{Published\\frontiers (30)} & \shortstack{Construction\\baselines (39)} & \shortstack{Valid\\fraction} & \shortstack{Gap\\closed} \\
\midrule
Qwen3.5-4B high & 7.55 & 0.07 [0.03, 0.11] & 0.08 [0.03, 0.15] & 0.38 & 0.000 \\
Qwen3.5-9B high & 14.52 & 0.12 [0.07, 0.17] & 0.17 [0.07, 0.27] & 0.43 & 0.000 \\
Qwen3.5-27B high & 16.18 & 0.13 [0.06, 0.22] & 0.18 [0.09, 0.29] & 0.52 & 0.000 \\
Qwen3.8-27B high & 38.07 & 0.31 [0.20, 0.42] & 0.44 [0.28, 0.59] & 0.62 & 0.004 \\
GPT-5.6 Luna medium & 16.70 & 0.16 [0.10, 0.24] & 0.17 [0.08, 0.27] & 0.57 & 0.000 \\
GPT-5.6 Luna high & 34.85 & 0.25 [0.16, 0.35] & 0.42 [0.29, 0.55] & 0.68 & 0.002 \\
DeepSeek V4.1 Flash low & 37.11 & 0.39 [0.29, 0.49] & 0.36 [0.23, 0.48] & 0.78 & 0.000 \\
DeepSeek V4.1 Flash high & 44.90 & 0.38 [0.26, 0.50] & 0.50 [0.34, 0.67] & 0.75 & 0.006 \\
Claude Fable 5.1 medium & 71.90 & 0.59 [0.48, 0.69] & 0.82 [0.65, 0.99] & 0.86 & 0.012 \\
Claude Fable 5.1 high & 43.73 & 0.54 [0.40, 0.68] & 0.36 [0.20, 0.51] & 0.52 & 0.003 \\
Claude Opus 5.5 medium & 66.70 & 0.58 [0.47, 0.68] & 0.74 [0.57, 0.93] & 0.88 & 0.006 \\
Claude Opus 5.5 high & 69.81 & 0.57 [0.46, 0.69] & 0.79 [0.61, 1.01] & 0.83 & 0.009 \\
GPT-6 Astra medium & 75.34 & 0.66 [0.58, 0.73] & 0.83 [0.71, 0.95] & 0.99 & 0.006 \\
GPT-6 Astra high & \textbf{91.90} & 0.76 [0.69, 0.83] & 1.04 [0.89, 1.22] & 1.00 & 0.016 \\
\midrule
GPT-6 Astra high + tools & 143.16 & 0.998 [0.995, 1.000] & 1.76 [1.43, 2.30] & 1.00 & 0.063 \\
GPT-5.6 Luna high + tools & 119.35 & 0.961 [0.931, 0.987] & 1.37 [1.09, 1.68] & 0.96 & 0.047 \\
Claude Opus 5.5 high + tools & \textbf{180.06} & 0.985 [0.965, 0.998] & 2.43 [1.52, 4.10] & 1.00 & 0.075 \\
\bottomrule
\end{tabular}

%% file: appendix/appendix.tex
\input{appendix/catalogue}

\input{appendix/cases}
\input{appendix/results}
\input{appendix/verification}
\input{appendix/references}
\input{appendix/parameters}
\input{appendix/settings}
\input{appendix/tools}
\endgroup

%% file: appendix/catalogue.tex
\begingroup
\raggedbottom
\setlength{\parskip}{4pt}
\hypersetup{colorlinks=true,linkcolor=appendixink,citecolor=appendixink,urlcolor=appendixink}
\section*{Appendix}
\begin{constructionbox}[title=Reading guide]
\small
\begin{tabularx}{\linewidth}{@{}lXr@{}}
\hyperref[app:families]{A\quad Task catalogue} & Fourteen families, examples and parameters controlling problem size & \pageref{app:families}\\
\hyperref[app:cases]{B\quad Representative constructions} & Four examples from object to score & \pageref{app:cases}\\
\hyperref[app:more]{C\quad Detailed results} & Model comparisons, size--quality curves and sensitivity & \pageref{app:more}\\
\hyperref[app:formats]{D\quad Verification} & Compact formats and algebraic certificates & \pageref{app:formats}\\
\hyperref[app:lit]{E\quad References and bounds} & Literature sources, construction baselines and the LABS target & \pageref{app:lit}\\
\hyperref[app:tiers]{F\quad Instance parameters} & Main and smaller-instance parameter settings & \pageref{app:tiers}\\
\hyperref[app:runner]{G\quad Evaluation settings} & Model settings and an example prompt & \pageref{app:runner}\\
\hyperref[app:tools]{H\quad Code and web access} & Tool-assisted results and token usage & \pageref{app:tools}
\end{tabularx}
\end{constructionbox}
\section{Task catalogue}
\label{app:families}
Each family page gives the objective, a checked small example, parameters controlling problem size and one evaluated instance with its reference values. Small diagrams illustrate the mathematical property. In the tables, the baseline is the better of an implemented mathematical construction and the ten-second search result. It can be weaker than a published frontier, which remains the scoring reference. The final column gives this baseline's relative quality, rather than a model score. Gap closed uses the scoring reference as its starting point, as defined in Section~\ref{sec:headroom}.
\begin{center}\small\setlength{\tabcolsep}{3pt}
\input{tables/families_table}
\captionof{table}{The fourteen families. R denotes search-resistant, S families that can benefit from search and U unclassified search sensitivity. These labels concern the reference search procedures studied here. Spherical codes and Heilbronn each contain two variants. Reference types may depend on the parameters.}
\label{tab:families}
\end{center}
\input{tables/family_catalogue}

%% file: tables/families_table.tex
\begin{tabularx}{\linewidth}{@{}P{.16\linewidth}YP{.12\linewidth}P{.09\linewidth}P{.22\linewidth}@{}}
\toprule
family & construction & objective & class & reference; bound or target \\
\midrule
\hyperref[family:capset]{cap set} & $S \subseteq \mathbb{F}_3^d$; no three collinear points (cap set) & $|S|$ & R & published; proven bound \\
\hyperref[family:spherical_code]{spherical codes} & vectors with exact coordinates; pairwise angle $\ge\theta$; $60^\circ$ or variable & count & R & mixed; proven bound \\
\hyperref[family:matmul]{matrix multiplication} & bilinear scheme for $(n\times m)$ by $(m\times p)$ matrices; computes the product exactly & rank $r$ (min) & S & published; proven bound \\
\hyperref[family:degdiam]{degree--diameter} & graph with max degree $d$; diameter $\le k$ & vertices & S & published; proven bound \\
\hyperref[family:covering]{covering} & $k$-subsets (blocks) of $[v]$; every $t$-subset lies in a block & blocks (min) & S & published; proven bound \\
\hyperref[family:schur]{Schur} & $k$-colouring of $\{1..N\}$; no monochromatic $x, y, x{+}y$ & $N$ & R & published; proven bound \\
\hyperref[family:apfree_q]{$\mathbb{F}_q^n$ AP-free} & $S \subseteq \mathbb{F}_q^n$, $q\in\{5,7\}$; no 3-term progression & $|S|$ & R & mixed; proven bound \\
\hyperref[family:mols]{MOLS} & Latin squares of order $n$, $n$ not a prime power; pairwise orthogonal & number of squares & R & published; proven bound \\
\midrule
\hyperref[family:labs]{LABS} & $\pm1$ sequence of length $N$; well-formed sequence & merit factor & S & construction; conjectured target \\
\hyperref[family:heilbronn]{Heilbronn} & $n$ points in a square or triangle; distinct points in the domain & min triangle area & S & construction; trivial bound \\
\hyperref[family:corners]{corners} & $S \subseteq [n]^2$; no corner & $|S|$ & R & construction; trivial bound \\
\hyperref[family:lineq]{linear-equation-free} & $S \subseteq \{1..n\}$; no non-trivial solution of the specified linear equation & $|S|$ & R & construction; trivial bound \\
\hyperref[family:shannon]{Shannon} & code in $\{0,\ldots,q-1\}^d$; every pair separates on the cycle & $|C|$ & R & construction; proven bound \\
\hyperref[family:trifference]{trifference} & ternary code of length $n$; every triple has at least $m$ coordinates with three distinct symbols & $|C|$ & U & construction; proven bound \\
\bottomrule
\end{tabularx}

%% file: tables/family_catalogue.tex
\appendixpage
\begin{minipage}{\linewidth}
\catalogueheading{capset}{Cap sets}
A large set must avoid every three-point arithmetic progression. \citep{edel2004capset,karapetyan2023caps}\par\smallskip
\begin{minipage}[c]{.25\linewidth}\centering
\includegraphics[width=.92\linewidth,height=73pt,keepaspectratio]{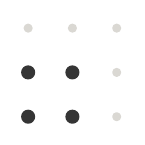}
\par{\scriptsize $d=2$: four points}\end{minipage}\hfill
\begin{minipage}[c]{.71\linewidth}
Maximise $|S|$ for $S\subseteq\mathbb F_3^d$ containing no three distinct points $x,y,z$ with $x+y+z=0$.\par\smallskip
\end{minipage}\par\smallskip
\textbf{Parameters and scaling.} Increasing $d$ enlarges the ambient space to $3^d$ points. Products lift finite caps to higher dimensions.
\begin{resultbox}[top=4pt,bottom=4pt]\small
\textbf{Evaluated example:} $d=10$.\par
\begin{tabularx}{\linewidth}{@{}XXXXX@{}}
Baseline & 10 s search & Reference & Bound or target & Baseline\textquotesingle s relative quality \\
2240 & 1140 & 2432 & 5619 & 0.921 \\
\end{tabularx}\par{\scriptsize Reference: published frontier; proven bound.}\par\smallskip
\end{resultbox}\end{minipage}
\par\vfill
\begin{minipage}{\linewidth}
\catalogueheading{corners}{Corner-free sets}
A dense grid set must avoid a prescribed three-point pattern. \citep{behrend1946}\par\smallskip
\begin{minipage}[c]{.25\linewidth}\centering
\includegraphics[width=.92\linewidth,height=73pt,keepaspectratio]{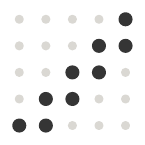}
\par{\scriptsize $n=5$: nine points}\end{minipage}\hfill
\begin{minipage}[c]{.71\linewidth}
Maximise $|S|$ for $S\subseteq\{0,\ldots,n-1\}^2$ containing no triple $(x,y),(x+t,y),(x,y+t)$ with $t\ne0$, including negative $t$.\par\smallskip
\end{minipage}\par\smallskip
\textbf{Parameters and scaling.} Increasing $n$ enlarges the grid. Progression-free difference sets supply constructions at many sizes.
\begin{resultbox}[top=4pt,bottom=4pt]\small
\textbf{Evaluated example:} $n=138$.\par
\begin{tabularx}{\linewidth}{@{}XXXXX@{}}
Baseline & 10 s search & Reference & Bound or target & Baseline\textquotesingle s relative quality \\
3120 & 2504 & 3120 & 19044 & 1.000 \\
\end{tabularx}\par{\scriptsize Reference: construction baseline; trivial bound.}\par\smallskip
\end{resultbox}\end{minipage}
\par\vfill
\appendixpage
\begin{minipage}{\linewidth}
\catalogueheading{schur}{Schur colourings}
No solution to $x+y=z$ may lie entirely within one colour class. \citep{rowley2021templates,bengone2026schur}\par\smallskip
\begin{minipage}[c]{.25\linewidth}\centering
\includegraphics[width=.92\linewidth,height=73pt,keepaspectratio]{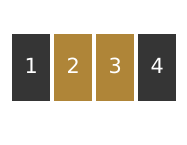}
\par{\scriptsize $k=2,\ N=4$}\end{minipage}\hfill
\begin{minipage}[c]{.71\linewidth}
Maximise $N$ such that $\{1,\ldots,N\}$ can be coloured with $k$ colours without a monochromatic solution of $x+y=z$. Solutions with $x=y$ are also forbidden.\par\smallskip
\end{minipage}\par\smallskip
\textbf{Parameters and scaling.} Increasing the number of colours $k$ gives new extremal targets. Recursive templates extend smaller colourings.
\begin{resultbox}[top=4pt,bottom=4pt]\small
\textbf{Evaluated example:} $k=7$.\par
\begin{tabularx}{\linewidth}{@{}XXXXX@{}}
Baseline & 10 s search & Reference & Bound or target & Baseline\textquotesingle s relative quality \\
1093 & 1093 & 1696 & 13699 & 0.644 \\
\end{tabularx}\par{\scriptsize Reference: published frontier; proven bound.}\par\smallskip
\end{resultbox}\end{minipage}
\par\vfill
\begin{minipage}{\linewidth}
\catalogueheading{apfree_q}{Progression-free sets over finite fields}
The ambient field changes which additive patterns a set must avoid. \citep{edel_caps,ellenberg2017capset}\par\smallskip
\begin{minipage}[c]{.25\linewidth}\centering
\includegraphics[width=.92\linewidth,height=73pt,keepaspectratio]{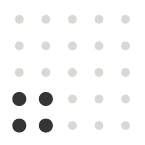}
\par{\scriptsize $q=5,\ n=2$}\end{minipage}\hfill
\begin{minipage}[c]{.71\linewidth}
Maximise $|S|$ for $S\subseteq\mathbb F_q^n$ containing no three distinct points satisfying $x+z=2y$.\par\smallskip
\end{minipage}\par\smallskip
\textbf{Parameters and scaling.} Varying $q$ or $n$ changes the task. Products and algebraic constructions give candidates across dimensions.
\begin{resultbox}[top=4pt,bottom=4pt]\small
\textbf{Evaluated example:} $q=5,\ n=6$.\par
\begin{tabularx}{\linewidth}{@{}XXXXX@{}}
Baseline & 10 s search & Reference & Bound or target & Baseline\textquotesingle s relative quality \\
360 & 360 & 649 & 7497 & 0.555 \\
\end{tabularx}\par{\scriptsize Reference: published frontier; proven bound.}\par\smallskip
\end{resultbox}\end{minipage}
\par\vfill
\appendixpage
\begin{minipage}{\linewidth}
\catalogueheading{spherical_code}{Spherical codes}
Directions must remain separated while as many vectors as possible are packed. \citep{zinoviev1999contact,ganzhinov2025lines,cohn2003kissing}\par\smallskip
\begin{minipage}[c]{.25\linewidth}\centering
\includegraphics[width=.92\linewidth,height=73pt,keepaspectratio]{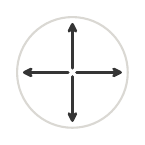}
\par{\scriptsize $d=2$: four directions}\end{minipage}\hfill
\begin{minipage}[c]{.71\linewidth}
\textbf{Fixed angle.} Maximise the number of nonzero vectors whose pairwise angles are at least $60^\circ$. After normalisation these give a kissing configuration. Coordinates are integers in $[-1000,1000]$; at $d=13$, we also allow $a+b\sqrt3$ with integer $|a|,|b|\le1000$.\par\smallskip
\textbf{Variable angle.} Maximise the number of nonzero vectors in $\mathbb Z^d$ satisfying $u\cdot v\le c\|u\|\|v\|$ for every distinct pair. Coordinates lie in $[-1000,1000]$, and the prescribed rational value $c=\cos\theta$ is checked exactly.\par\smallskip
\end{minipage}\par\smallskip
\textbf{Parameters and scaling.} Dimension $d$ and the angle threshold control the packing problem. Lattices, orbits and signed supports supply constructions.
\begin{resultbox}[top=4pt,bottom=4pt]\small
\textbf{Evaluated example:} $d=14$.\par
\begin{tabularx}{\linewidth}{@{}XXXXX@{}}
Baseline & 10 s search & Reference & Bound or target & Baseline\textquotesingle s relative quality \\
1932 & 876 & 1932 & 3174 & 1.000 \\
\end{tabularx}\par{\scriptsize Reference: published frontier; proven bound.}\par\smallskip
\textbf{Evaluated example:} $d=12,\ c=24/100$.\par
\begin{tabularx}{\linewidth}{@{}XXXXX@{}}
Baseline & 10 s search & Reference & Bound or target & Baseline\textquotesingle s relative quality \\
28 & 24 & 28 & 1431 & 1.000 \\
\end{tabularx}\par{\scriptsize Reference: construction baseline; proven bound.}\par\smallskip
\end{resultbox}\end{minipage}
\par\vfill
\begin{minipage}{\linewidth}
\catalogueheading{labs}{Low-autocorrelation binary sequences}
A binary sequence should have low autocorrelation at nonzero shifts. \citep{packebusch2016labs}\par\smallskip
\begin{minipage}[c]{.25\linewidth}\centering
\includegraphics[width=.92\linewidth,height=73pt,keepaspectratio]{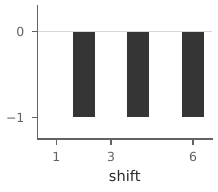}
\par{\scriptsize $N=7$: autocorrelation}\end{minipage}\hfill
\begin{minipage}[c]{.71\linewidth}
Choose $s\in\{-1,1\}^N$ to maximise the merit factor $N^2/(2\sum_{k=1}^{N-1}C_k^2)$, where $C_k=\sum_{i=1}^{N-k}s_i s_{i+k}$ is the aperiodic autocorrelation.\par\smallskip
\end{minipage}\par\smallskip
\textbf{Parameters and scaling.} Increasing $N$ tests whether a construction retains a high merit factor at longer lengths.
\begin{resultbox}[top=4pt,bottom=4pt]\small
\textbf{Evaluated example:} $N=280$.\par
\begin{tabularx}{\linewidth}{@{}XXXXX@{}}
Baseline & 10 s search & Reference & Bound or target & Baseline\textquotesingle s relative quality \\
5.57769 & 4.4708 & 5.57769 & 12.32 & 1.000 \\
\end{tabularx}\par{\scriptsize Reference: construction baseline; conjectured target.}\par\smallskip
\end{resultbox}\end{minipage}
\par\vfill
\appendixpage
\begin{minipage}{\linewidth}
\catalogueheading{heilbronn}{Heilbronn triangles}
Place the points so that the smallest triangle area is as large as possible. (definitions and references in Appendix~\ref{app:lit})\par\smallskip
\begin{minipage}[c]{.25\linewidth}\centering
\includegraphics[width=.92\linewidth,height=73pt,keepaspectratio]{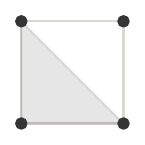}
\par{\scriptsize $n=4$: minimum area $1/2$}\end{minipage}\hfill
\begin{minipage}[c]{.71\linewidth}
\textbf{Square.} Place $n$ distinct points in $[0,1]^2$ to maximise the smallest area of a triangle determined by three points. Coordinates are multiples of $10^{-4}$.\par\smallskip
\textbf{Triangle.} Place $n$ distinct integer-grid points in a specified triangle $T$ to maximise the smallest triangle area, divided by the area of $T$.\par\smallskip
\end{minipage}\par\smallskip
\textbf{Parameters and scaling.} Increasing $n$ introduces more triangle constraints. The square and triangular domains are evaluated separately.
\begin{resultbox}[top=4pt,bottom=4pt]\small
\textbf{Evaluated example:} $n=46$.\par
\begin{tabularx}{\linewidth}{@{}XXXXX@{}}
Baseline & 10 s search & Reference & Bound or target & Baseline\textquotesingle s relative quality \\
0.00047375 & 0.00047375 & 0.00047375 & 0.0227273 & 1.000 \\
\end{tabularx}\par{\scriptsize Reference: construction baseline; trivial bound.}\par\smallskip
\textbf{Evaluated example:} $n=32$, triangular domain (Table~\ref{tab:triangle_domains}).\par
\begin{tabularx}{\linewidth}{@{}XXXXX@{}}
Baseline & 10 s search & Reference & Bound or target & Baseline\textquotesingle s relative quality \\
0.0014885 & 0.0014885 & 0.0014885 & 0.0333333 & 1.000 \\
\end{tabularx}\par{\scriptsize Reference: construction baseline; trivial bound.}\par\smallskip
\end{resultbox}\end{minipage}
\par\vfill
\begin{minipage}{\linewidth}
\catalogueheading{degdiam}{Degree--diameter graphs}
A sparse graph must keep all vertices close to one another. \citep{miller2013degdiam,comellas_degdiam}\par\smallskip
\begin{minipage}[c]{.25\linewidth}\centering
\includegraphics[width=.92\linewidth,height=73pt,keepaspectratio]{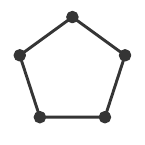}
\par{\scriptsize $d=2,\ k=2,\ N=5$}\end{minipage}\hfill
\begin{minipage}[c]{.71\linewidth}
Maximise the number of vertices of a simple connected undirected graph with maximum degree at most $d$ and diameter at most $k$.\par\smallskip
\end{minipage}\par\smallskip
\textbf{Parameters and scaling.} Varying degree $d$ and diameter $k$ generates new graph problems. Larger diameters test the growth of a construction.
\begin{resultbox}[top=4pt,bottom=4pt]\small
\textbf{Evaluated example:} $d=4,\ k=5$.\par
\begin{tabularx}{\linewidth}{@{}XXXXX@{}}
Baseline & 10 s search & Reference & Bound or target & Baseline\textquotesingle s relative quality \\
61 & 61 & 364 & 485 & 0.168 \\
\end{tabularx}\par{\scriptsize Reference: published frontier; proven bound.}\par\smallskip
\end{resultbox}\end{minipage}
\par\vfill
\appendixpage
\begin{minipage}{\linewidth}
\catalogueheading{lineq}{Linear-equation-free sets}
A large integer set must avoid a specified linear relation. \citep{behrend1946,ruzsa1993lineq}\par\smallskip
\begin{minipage}[c]{.25\linewidth}\centering
\includegraphics[width=.92\linewidth,height=73pt,keepaspectratio]{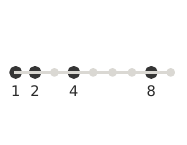}
\par{\scriptsize $a=b=1,\ n=9$}\end{minipage}\hfill
\begin{minipage}[c]{.71\linewidth}
Maximise $|S|$ for $S\subseteq\{1,\ldots,n\}$ such that $ax+by=(a+b)z$ has no solution in $S$ except $x=y=z$.\par\smallskip
\end{minipage}\par\smallskip
\textbf{Parameters and scaling.} The coefficients $a,b$ select the relation; $n$ sets the interval size. Digit constructions can extend to larger intervals.
\begin{resultbox}[top=4pt,bottom=4pt]\small
\textbf{Evaluated example:} $a=3,\ b=4,\ n=19044$.\par
\begin{tabularx}{\linewidth}{@{}XXXXX@{}}
Baseline & 10 s search & Reference & Bound or target & Baseline\textquotesingle s relative quality \\
630 & 630 & 630 & 19044 & 1.000 \\
\end{tabularx}\par{\scriptsize Reference: construction baseline; trivial bound.}\par\smallskip
\end{resultbox}\end{minipage}
\par\vfill
\begin{minipage}{\linewidth}
\catalogueheading{covering}{Covering designs}
A small collection of blocks must cover every required subset. \citep{gordon1995covering,ljcr}\par\smallskip
\begin{minipage}[c]{.25\linewidth}\centering
\includegraphics[width=.92\linewidth,height=73pt,keepaspectratio]{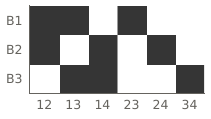}
\par{\scriptsize Three blocks cover six pairs}\end{minipage}\hfill
\begin{minipage}[c]{.71\linewidth}
Minimise the number of $k$-element blocks drawn from a $v$-element set such that every $t$-element subset is contained in at least one block.\par\smallskip
\end{minipage}\par\smallskip
\textbf{Parameters and scaling.} The parameters $(v,k,t)$ control the universe, block size and subsets to cover. Larger settings test covering efficiency.
\begin{resultbox}[top=4pt,bottom=4pt]\small
\textbf{Evaluated example:} $v=18,\ k=7,\ t=4$.\par
\begin{tabularx}{\linewidth}{@{}XXXXX@{}}
Baseline & 10 s search & Reference & Bound or target & Baseline\textquotesingle s relative quality \\
179 & 179 & 126 & 111 & 0.704 \\
\end{tabularx}\par{\scriptsize Reference: published frontier; proven bound.}\par\smallskip
\end{resultbox}\end{minipage}
\par\vfill
\appendixpage
\begin{minipage}{\linewidth}
\catalogueheading{matmul}{Matrix multiplication}
A bilinear construction reduces the number of scalar multiplications. \citep{strassen1969,fawzi2022alphatensor}\par\smallskip
\begin{minipage}[c]{.25\linewidth}\centering
\includegraphics[width=.92\linewidth,height=73pt,keepaspectratio]{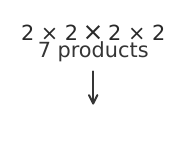}
\par{\scriptsize Strassen: seven products}\end{minipage}\hfill
\begin{minipage}[c]{.71\linewidth}
Minimise the number $r$ of scalar multiplications in a bilinear algorithm for $A\in\mathbb R^{n\times m}$ and $B\in\mathbb R^{m\times p}$. The identity $AB=\sum_{j=1}^r (u_j\cdot A)(v_j\cdot B)W_j$ must hold for all inputs, with each dot product taken over matrix entries. Coefficients are integers in $[-4,4]$.\par\smallskip
\end{minipage}\par\smallskip
\textbf{Parameters and scaling.} Changing the matrix dimensions $(n,m,p)$ defines new multiplication problems. Block composition extends schemes to larger products.
\begin{resultbox}[top=4pt,bottom=4pt]\small
\textbf{Evaluated example:} $n=4,\ m=4,\ p=5$.\par
\begin{tabularx}{\linewidth}{@{}XXXXX@{}}
Baseline & 10 s search & Reference & Bound or target & Baseline\textquotesingle s relative quality \\
72 & 72 & 61 & 20 & 0.847 \\
\end{tabularx}\par{\scriptsize Reference: published frontier; proven bound.}\par\smallskip
\end{resultbox}\end{minipage}
\par\vfill
\begin{minipage}{\linewidth}
\catalogueheading{mols}{Mutually orthogonal Latin squares}
When any two squares are superimposed, every ordered pair of symbols must occur exactly once. \citep{miller2024mols,abel2015mols}\par\smallskip
\begin{minipage}[c]{.25\linewidth}\centering
\includegraphics[width=.92\linewidth,height=73pt,keepaspectratio]{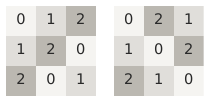}
\par{\scriptsize Two squares of order three}\end{minipage}\hfill
\begin{minipage}[c]{.71\linewidth}
Find as many mutually orthogonal Latin squares of order $n$ as possible. Each symbol appears once in every row and column; superimposing any two squares must give all $n^2$ ordered symbol pairs exactly once.\par\smallskip
\end{minipage}\par\smallskip
\textbf{Parameters and scaling.} Changing the order $n$ changes the design problem. Algebraic constructions and combinations of smaller designs supply candidates.
\begin{resultbox}[top=4pt,bottom=4pt]\small
\textbf{Evaluated example:} $n=14$.\par
\begin{tabularx}{\linewidth}{@{}XXXXX@{}}
Baseline & 10 s search & Reference & Bound or target & Baseline\textquotesingle s relative quality \\
4 & 1 & 4 & 12 & 1.000 \\
\end{tabularx}\par{\scriptsize Reference: published frontier; proven bound.}\par\smallskip
\end{resultbox}\end{minipage}
\par\vfill
\appendixpage
\begin{minipage}{\linewidth}
\catalogueheading{shannon}{Shannon codes}
In the cycle model, adjacent symbols can be confused; every pair of codewords must have a coordinate that distinguishes them. \citep{lovasz1979shannon,polak2019shannon}\par\smallskip
\begin{minipage}[c]{.25\linewidth}\centering
\includegraphics[width=.92\linewidth,height=73pt,keepaspectratio]{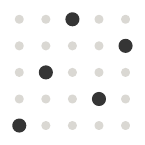}
\par{\scriptsize $C_5^2$: five codewords}\end{minipage}\hfill
\begin{minipage}[c]{.71\linewidth}
Maximise the size of a code $C\subseteq\mathbb Z_q^d$ such that every two distinct words differ by cyclic distance at least two in some coordinate. Equivalently, $C$ is an independent set in the $d$-fold strong power of the cycle $C_q$.\par\smallskip
\end{minipage}\par\smallskip
\textbf{Parameters and scaling.} The cycle size $q$ and codeword length $d$ define the task. Finite factors give short certificates for very large products.
\begin{resultbox}[top=4pt,bottom=4pt]\small
\textbf{Evaluated example:} $q=7,\ d=24$.\par
\begin{tabularx}{\linewidth}{@{}XXXXX@{}}
Baseline & 10 s search & Reference & Bound or target & Baseline\textquotesingle s relative quality \\
$1.9411\times10^{12}$ & $1.9411\times10^{12}$ & $1.9411\times10^{12}$ & $3.16219\times10^{12}$ & 1.000 \\
\end{tabularx}\par{\scriptsize Reference: construction baseline; proven bound.}\par\smallskip
\end{resultbox}\end{minipage}
\par\vfill
\begin{minipage}{\linewidth}
\catalogueheading{trifference}{Trifference codes}
Every triple of distinct codewords must display all three symbols in at least $m$ coordinates. \citep{bishnoi2024trifferent,bishnoi2025generalized}\par\smallskip
\begin{minipage}[c]{.25\linewidth}\centering
\includegraphics[width=.92\linewidth,height=73pt,keepaspectratio]{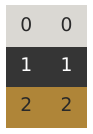}
\par{\scriptsize $n=2,\ m=2$: three words}\end{minipage}\hfill
\begin{minipage}[c]{.71\linewidth}
Maximise the size of a ternary code $C\subseteq\{0,1,2\}^n$ such that, for every three distinct words, at least $m$ coordinates contain all three symbols.\par\smallskip
\end{minipage}\par\smallskip
\textbf{Parameters and scaling.} Length $n$ and separation multiplicity $m$ define the task. Linear and concatenated codes admit compact certificates.
\begin{resultbox}[top=4pt,bottom=4pt]\small
\textbf{Evaluated example:} $n=64,\ m=1$.\par
\begin{tabularx}{\linewidth}{@{}XXXXX@{}}
Baseline & 10 s search & Reference & Bound or target & Baseline\textquotesingle s relative quality \\
19683 & 19683 & 19683 & $3.72281\times10^{11}$ & 1.000 \\
\end{tabularx}\par{\scriptsize Reference: construction baseline; proven bound.}\par\smallskip
\end{resultbox}\end{minipage}
\par\vfill

%% file: appendix/cases.tex
\appendixpage
\section{Representative constructions}
\label{app:cases}
\input{tables/appendix_case_numbers}
\subsection{Cap sets: a small ingredient at twice the dimension}
\label{case:capset}
\textbf{Model:} GPT-6 Astra, high effort, tool-free. \textbf{Instance:} $d=12$, problem-size experiment.

\begin{taskbox}[title=Task]
Construct a subset of $\mathbb F_3^{12}$ with no three distinct points summing to zero. The objective is the number of points. The published frontier contains 12,928 points.
\end{taskbox}

\begin{center}\includegraphics[width=\linewidth]{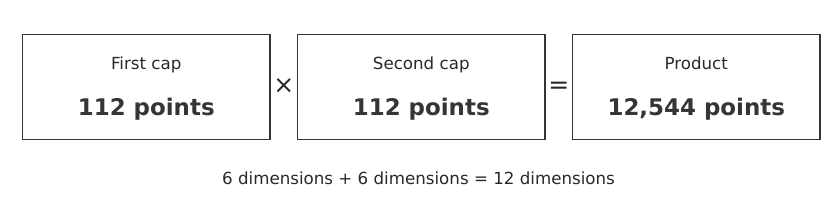}\end{center}

\begin{constructionbox}[title=Construction and verification]
The answer takes the Cartesian product of two copies of a 112-point cap $C\subseteq\mathbb F_3^6$:
\[
S=C\times C,\qquad |S|=112^2=12544.
\]
Each finite factor is checked for forbidden triples. In characteristic three, a zero-sum triple in a cap has all three entries equal: three distinct entries are forbidden, and two equal entries force the third to be equal. Applying this fact in each factor shows that any zero-sum triple in $C\times C$ consists of one repeated point.

Both answers at this problem size use this product. The construction makes the dependence on dimension explicit: repeating a factor adds dimensions and multiplies cardinalities.
\end{constructionbox}

\begin{resultbox}[title=Measured quality]
\begin{tabularx}{\linewidth}{@{}XXX@{}}
Model construction & Published frontier & Relative quality \\
\textbf{12,544 points} & 12,928 points & $12544/12928=\mathbf{\CaseCapRatio}$
\end{tabularx}
\end{resultbox}

The product describes a large, valid cap set close to the published frontier using only its smaller factors. The size--quality curves in Appendix~\ref{app:ladder_results} compare how this construction and graph constructions behave as their parameters grow.

\appendixpage
\subsection{Shannon codes: better factors produce a larger code}
\label{case:shannon}
\textbf{Model:} Qwen3.8-27B, high effort, tool-free. \textbf{Instance:} length 24 over seven symbols ($C_7^{24}$).

\begin{taskbox}[title=Task]
Construct codewords of length 24 over $\mathbb Z_7$. Every pair must differ by cyclic distance at least two in some coordinate. The objective is the number of codewords.
\end{taskbox}

\begin{center}\includegraphics[width=\linewidth]{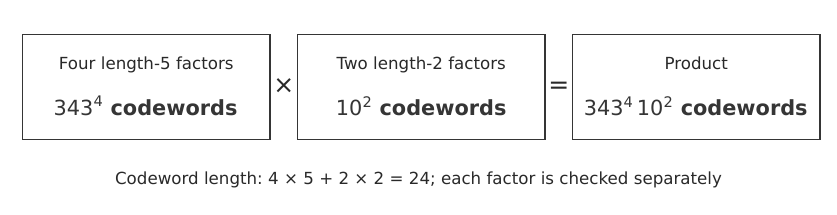}\end{center}

\begin{constructionbox}[title=Construction and verification]
The submitted factors contain 343 words of length five and ten words of length two. Their multiplicities are four and two:
\[
4\cdot5+2\cdot2=24,\qquad M=343^4\,10^2=1{,}384{,}128{,}720{,}100.
\]
The evaluator verifies the finite factors and computes the exact product size. Concatenating words from valid Shannon-code factors preserves the separation condition~\citep{polak2019shannon,tandon2026shannon}: two distinct product words differ in at least one factor, which supplies the coordinate required for separation.

GPT-6 Astra at medium effort and DeepSeek V4.1 Flash at low effort instead return $10^{12}$ codewords on this instance. Qwen3.8-27B's choice of factors produces a larger valid code at the same length.
\end{constructionbox}

\begin{resultbox}[title=Measured quality]
Construction baseline: \CaseShannonReference{} words.\\
Qwen3.8-27B relative quality: \textbf{\CaseShannonRatio}.\\
Size advantage over the two $10^{12}$-word answers: \textbf{\CaseShannonGain\%}.
\end{resultbox}

The comparison measures the quality of the finite ingredients and their composition. Even for codes with more than a trillion words, verification gives their exact sizes and makes the improvement measurable.

\appendixpage
\subsection{Trifference: ten rows describe 59,049 codewords}
\label{case:trifference}
\textbf{Model:} GPT-6 Astra, high effort, tool-free. \textbf{Instance:} $n=64$, $m=1$.

\begin{taskbox}[title=Task]
Construct a ternary code in which every three distinct words have at least one coordinate containing all three symbols. The construction baseline has 19,683 words.
\end{taskbox}

\begin{center}\includegraphics[width=\linewidth]{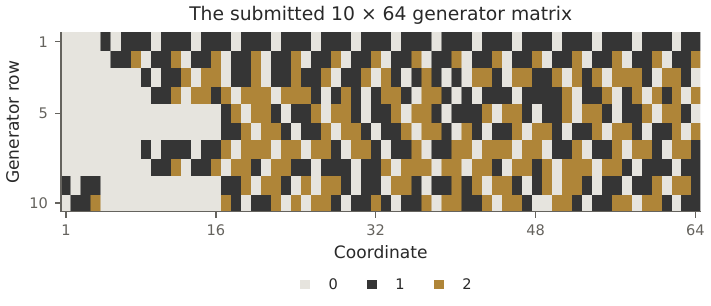}\end{center}

\begin{constructionbox}[title=Construction and verification]
The answer is a $10\times64$ ternary generator matrix $G$. Its rows span a linear code
\[
C=\{uG:u\in\mathbb F_3^{10}\},\qquad \operatorname{rank}(G)=10,
\qquad |C|=3^{10}=59049.
\]
The coloured matrix above is the submitted generator: grey, black and gold denote 0, 1 and 2. The hyperplane test in Appendix~\ref{app:trifference_certificates} checks the separation condition without listing every triple of codewords. Both verifiers accepted the construction.
\end{constructionbox}

\begin{resultbox}[title=Measured quality]
\begin{tabularx}{\linewidth}{@{}XXX@{}}
Model construction & Construction baseline & Relative quality \\
\textbf{59,049 words} & 19,683 words & $59049/19683=\mathbf{3}$
\end{tabularx}
\end{resultbox}

A short certificate describes a code three times as large as the construction baseline, giving relative quality 3. The other three evaluated lengths test whether that quality extends further.

\appendixpage
\subsection{Schur colourings: successive advances on one scale}
\label{case:schur}
\textbf{Source:} published constructions. \textbf{Instance:} eight colours.

\begin{taskbox}[title=Task]
Colour $\{1,\ldots,N\}$ with eight colours so that no colour class contains a solution of $x+y=z$, including $x=y$. Larger $N$ gives a stronger construction.
\end{taskbox}

\begin{center}\includegraphics[width=\linewidth]{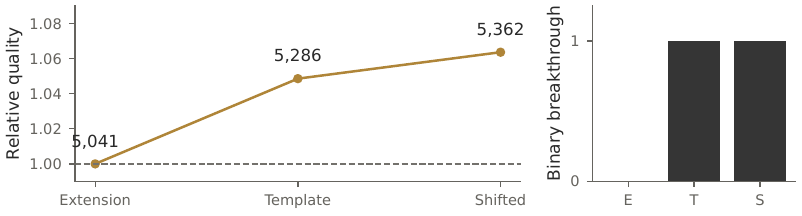}\end{center}

\begin{constructionbox}[title=Published constructions]
Recursive extension of the 1,680-element seven-colour construction reported by \citet{fredricksen2000schur} gives the historical reference
\[
N_0=3\cdot1680+1=5041.
\]
The template in \citet{rowley2021templates} gives $33\cdot160+6=5286$. Shifted templates give $10\cdot536+2=5362$~\citep{bengone2026schur}. We compare the published constructions against the same historical reference $N_0$.
\end{constructionbox}

\begin{resultbox}[title=Measured progress]
\centering\footnotesize\setlength{\tabcolsep}{5pt}
\input{tables/historical_progress}
\captionof{table}{Selected published Schur constructions with eight colours, scored against the historical reference $N_0=5041$. The last column records whether the construction exceeds this reference. Citations identify the accounts used for each construction.}
\label{tab:history}
\end{resultbox}

Both advances count as one binary success, while relative quality preserves their different magnitudes. The current benchmark uses 5,362 as the eight-colour reference and extends the challenge by increasing the number of colours under the same construction rule.

%% file: tables/appendix_case_numbers.tex
\newcommand{\CaseCapRatio}{0.970}
\newcommand{\CaseShannonReference}{1,941,100,559,147}
\newcommand{\CaseShannonRatio}{0.713}
\newcommand{\CaseShannonGain}{38.4}

%% file: tables/historical_progress.tex
\begin{tabular}{@{}lrrr@{}}
\toprule
Published construction & $N$ & $N/5041$ & $\mathbf{1}\{N>5041\}$ \\
\midrule
Recursive extension~\citep{fredricksen2000schur} & 5,041 & 1.000 & 0 \\
Rowley template~\citep{rowley2021templates} & 5,286 & 1.049 & 1 \\
Shifted template~\citep{bengone2026schur} & 5,362 & 1.064 & 1 \\
\bottomrule
\end{tabular}

%% file: appendix/results.tex
\appendixpage
\section{Detailed experimental results}
\label{app:more}
\subsection{Quality across the tool-free model comparison}
\readingnote{Although every tool-free configuration has a zero breakthrough rate on the 30 published-frontier instances, their mean relative quality spans \POneMin--\POneMax. The two panels place this comparison alongside quality on the 39 construction-baseline instances; Appendix~\ref{app:tools} extends the evaluation to tool-assisted responses.}
\begin{center}
\centering
\includegraphics[width=\linewidth]{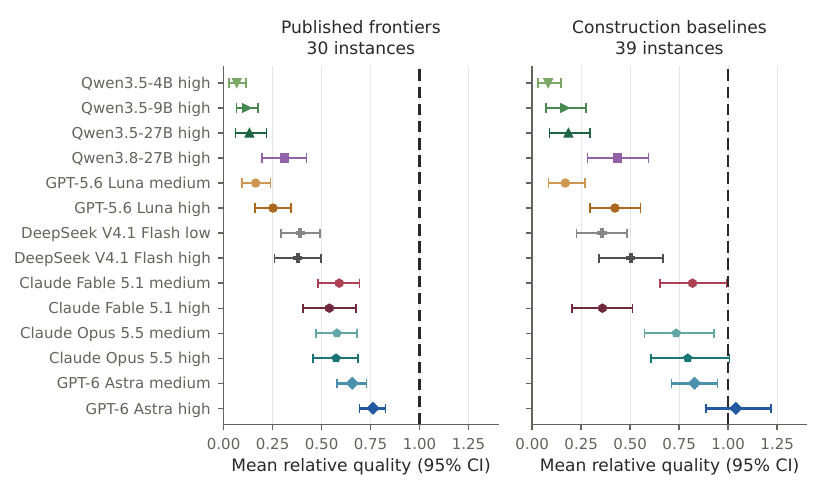}
\captionof{figure}{Fourteen-family construction quality for all fourteen tool-free configurations. Whiskers show $95\%$ confidence intervals obtained by resampling instances within each reference group: 30 with published frontiers and 39 with construction baselines.}
\label{fig:headline}
\end{center}
The overall score in Table~\ref{tab:headline} combines all 69 instances using the definition in Section~\ref{sec:scoring}.
\appendixpage
\subsection{Performance across parameter ranges}
\readingnote{The non-code experiments use three parameter ranges: A1 contains small familiar instances, A2 uses larger settings, and A3 supplies the main evaluation. Family coverage varies across these groups. Appendix~\ref{app:ladder_results} instead varies size within each family, allowing construction quality to be followed as the task grows.}
\begin{center}
\includegraphics[width=\linewidth]{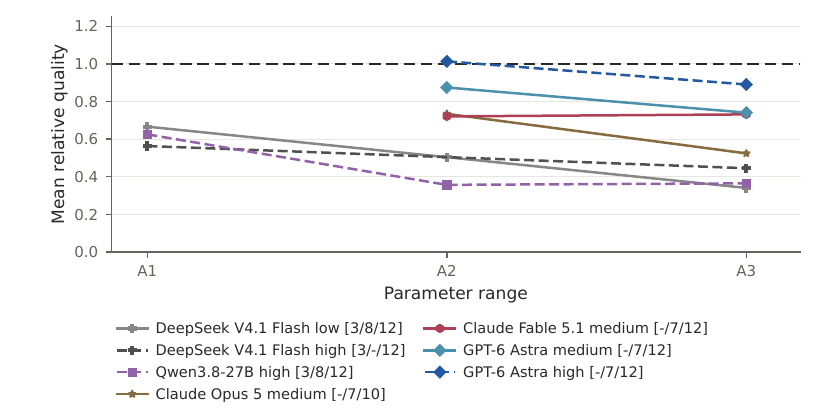}
\captionof{figure}{Performance across parameter ranges. Legend brackets give the numbers of evaluated families in A1, A2 and A3, respectively; a dash denotes an unmeasured group. The family sets differ across groups. Claude Opus 5 is an earlier checkpoint included in these auxiliary comparisons.}
\label{fig:tiers}
\end{center}
\appendixpage
\subsection{Model size and reasoning effort}
\label{app:effort_results}
\readingnote{Qwen3.5's mean quality and validity increase across the three checkpoint sizes. The GPT-6 Astra experiment extends the reasoning-effort comparison to xhigh on shared smaller instances.}
On the main 69 instances, Astra medium and high return \AstraMedValidCalls{} and \AstraHighValidCalls{} valid constructions. To include xhigh effort, we also compare Astra on \AstraEffN{} shared smaller instances. Its mean relative quality is $\AstraMedATwoHFR$ at medium, $\AstraHighATwoHFR$ at high and $\AstraXhighATwoHFR$ at xhigh.
\begin{center}
\centering
\includegraphics[width=\linewidth]{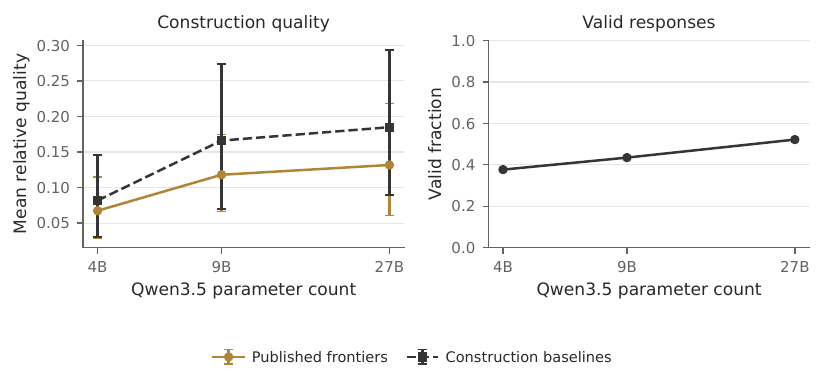}
\captionof{figure}{Model-size comparison within Qwen3.5: 4B, 9B and 27B at high effort on the 69 instances. Left: mean relative quality with bootstrap CIs. Right: valid-response fraction. Model size varies while the problems remain fixed.}
\label{fig:model_size}
\end{center}
\begin{center}
\centering
\includegraphics[width=\linewidth]{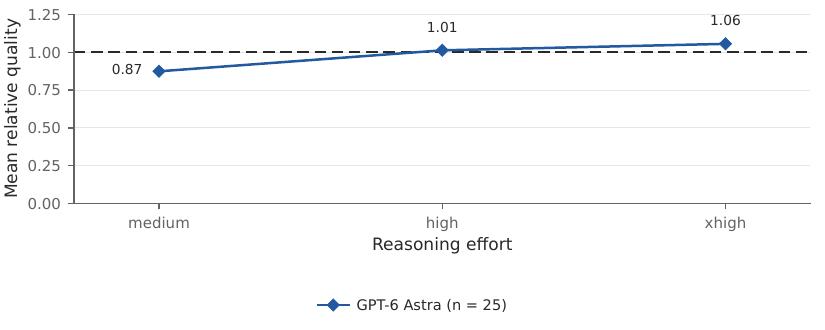}
\captionof{figure}{Mean relative quality for GPT-6 Astra at medium, high and xhigh effort on the same \AstraEffN{} smaller instances (A2). Multiple responses are averaged within each instance. All responses are valid at each effort level.}
\label{fig:effort}
\end{center}
\appendixpage
\subsection{Size--quality curves: cap sets, spherical codes and graphs}
\label{app:ladder_results}
\readingnote{Cap-set products allow GPT-6 Astra to retain quality at larger dimensions. Its graphs with maximum degree three remain valid while falling behind the reference as diameter increases. The lower panels apply the displayed transformations to construction sizes, showing how the model answers and references scale.}
\begin{center}\includegraphics[width=\linewidth]{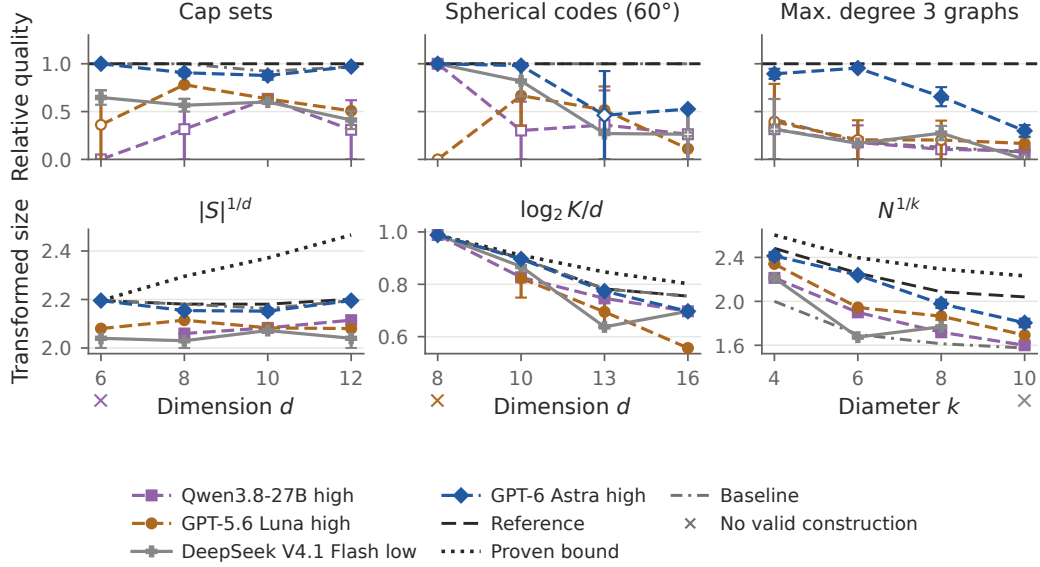}
\captionof{figure}{Size--quality curves for cap sets, spherical codes and degree--diameter graphs. Upper panels average two independent responses per size, including zero scores; whiskers show their range and hollow markers indicate at least one invalid response. Lower panels use valid responses only; crosses below the axes mark sizes with no valid construction, using the model's colour. Transformations and line conventions follow Figure~\ref{fig:ladder}.}
\label{fig:ladder_first}
\end{center}
\appendixpage
\subsection{Size--quality curves: colourings, coverings and integer sets}
\readingnote{These curves vary the number of Schur colours, the size of the set covered by a design, and the integer interval from which a progression-free subset is chosen. The integer task supplements the five benchmark families in this experiment. The lower panels show the transformed construction sizes specified above each panel.}
\begin{center}\includegraphics[width=\linewidth]{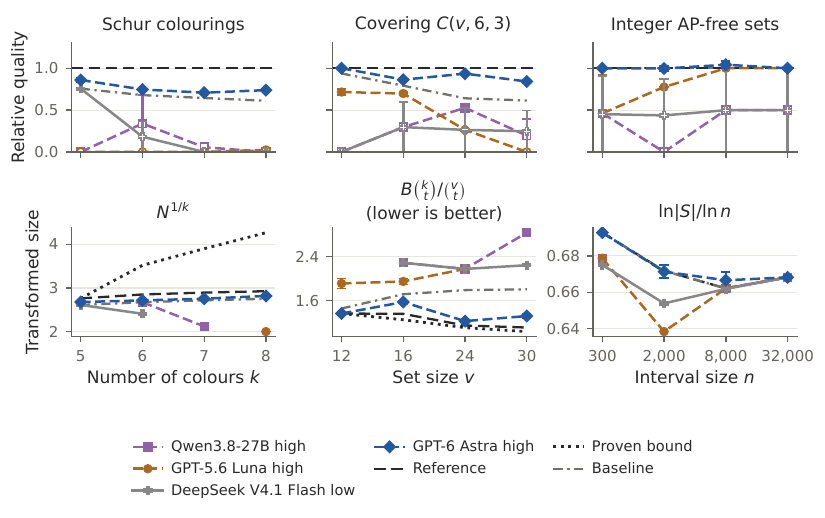}
\captionof{figure}{Size--quality curves for Schur colourings, covering designs and integer progression-free sets. Each size has two independent responses per configuration. Upper panels include zero scores; lower panels use valid answers. For Schur, $N$ is the largest coloured integer and $k$ the number of colours. For coverings, $B$ blocks of size $k$ cover every $t$-subset of a $v$-element set. For integer sets, $|S|$ elements are selected from $\{1,\ldots,n\}$. Whiskers show the range of the responses included in each panel.}
\label{fig:ladder_designs}
\end{center}
\appendixpage
\subsection{A changing reference: graphs with maximum degree four}
\readingnote{GPT-6 Astra's relative quality is \QuarticMTwo{} at diameter 5, $0.98$ at diameter 6 and \QuarticMFour{} at diameter 7. Between diameters 5 and 7, its graph size changes from 192 to 195 vertices, while the reference grows more quickly.}
\begin{center}\includegraphics[width=\linewidth]{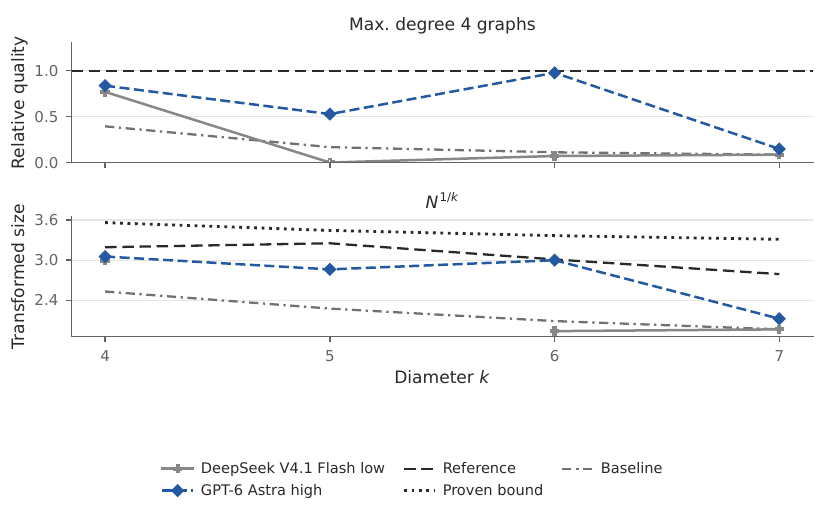}
\captionof{figure}{Graphs with maximum degree four at diameters 4--7. The lower panel plots $N^{1/k}$, where $N$ is the number of vertices and $k$ the allowed diameter. Each point represents one response from GPT-6 Astra at high effort or DeepSeek V4.1 Flash at low effort.}
\label{fig:ladder_quartic}
\end{center}
\appendixpage
\subsection{Problem-size experiments: instances and references}
\small
\input{tables/ladder_parameters}
\normalsize
\appendixpage
\subsection{Problem-size experiments: relative quality}
\small\setlength{\tabcolsep}{2pt}
\input{tables/ladder_scores}
\normalsize
\appendixpage
\subsection{Code quality across increasing lengths}
\readingnote{The same finite factors can generate large Shannon products, while trifference certificates represent codes far beyond an explicit list. Relative quality measures their quality at each length. The lower panels express code size $M$ as a per-coordinate growth base $M^{1/d}$ for Shannon and a ternary rate $\log_3 M/n$ for trifference, where $d$ and $n$ are codeword lengths.}
\begin{center}
\centering
\includegraphics[width=\linewidth]{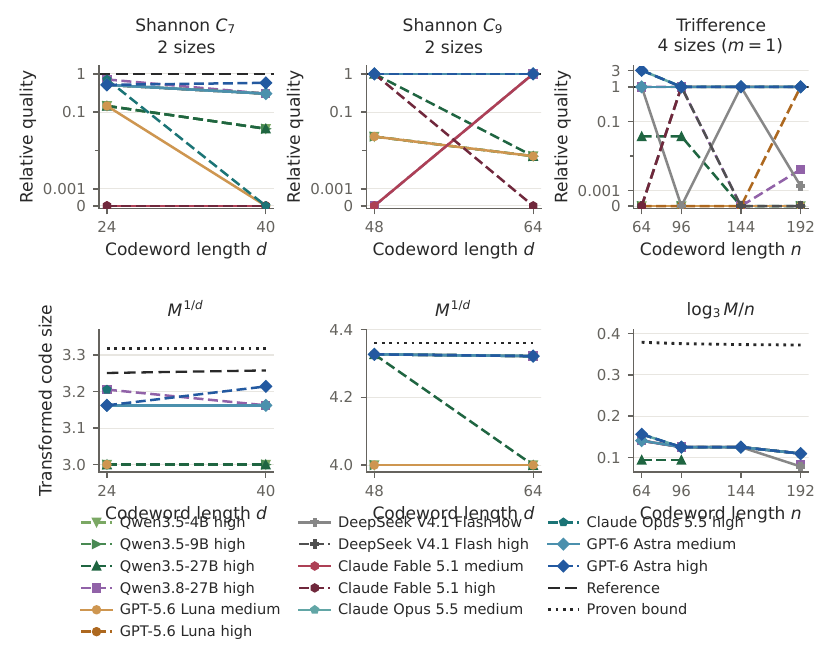}
\captionof{figure}{Code-family size--quality curves for all fourteen configurations. Top: relative quality, including zero scores. Bottom: per-coordinate growth for Shannon and ternary rate for trifference, using valid answers only; missing points break the curves. Dark dashed lines are references; dark dotted lines are proven upper bounds. Shannon has two codeword lengths per cycle size; trifference has four lengths at $m=1$. Each point represents one response.}
\label{fig:code_scale}
\end{center}
\appendixpage
\subsection{Code-family results}
\label{app:code_comparison}
Table~\ref{tab:code_comparison} compares all fourteen tool-free configurations on four Shannon and four trifference instances. Appendix~\ref{case:shannon} gives the Shannon product construction and its verification.
\begin{center}\footnotesize\setlength{\tabcolsep}{3pt}
\input{tables/code_multimodel}
\captionof{table}{Tool-free code-family results on four instances per family, including zero scores. A ratio of $1$ matches the construction baseline. The last column gives the trifference code size at length 192, or zero for failure.}
\label{tab:code_comparison}
\end{center}

Without tools, GPT-6 Astra at high effort and Claude Opus 5.5 at medium and high effort each produce 59,049 trifference codewords at length 64, three times the baseline. All three configurations match the baselines at the other three lengths. Astra medium and Claude Fable 5.1 medium match the baselines at all four lengths, certifying $3^{21}$ codewords at length 192. GPT-5.6 Luna at medium effort has no valid trifference answer, while Luna high matches the baseline only at length 192. Figure~\ref{fig:code_scale} shows how code size scales with length and compares the results with the references and proven bounds.

\appendixpage
Tables~\ref{tab:code_multimodel_detail} and \ref{tab:code_detail_trifference} give the individual outcomes behind these averages.
\begin{center}\small\setlength{\tabcolsep}{5pt}
\input{tables/code_detail_shannon}
\captionof{table}{Shannon relative quality on the four evaluated instances. A dagger marks an invalid or budget-exhausted response, scored zero.}
\label{tab:code_multimodel_detail}
\end{center}
\begin{center}\small\setlength{\tabcolsep}{5pt}
\input{tables/code_detail_trifference}
\captionof{table}{Trifference relative quality at separation multiplicity $m=1$. A dagger marks an invalid or budget-exhausted response, scored zero.}
\label{tab:code_detail_trifference}
\end{center}
\appendixpage
\subsection{Failures and answer formats}
Zero-score responses arise from invalid mathematical constructions, malformed descriptions, or empty and budget-exhausted answers, with the reason for each response recorded in the released data. A format error can obscure a valid construction, as in DeepSeek V4.1 Flash's length-96 trifference response at low effort, which supplies generator rows as integer arrays rather than the required digit strings. Expressing those rows in the required format gives a valid 19,683-word code, although the reported score uses the submitted representation.

\subsection{Smaller trifference instances}
\label{app:codes}
Four smaller trifference instances test the quality of finite codes that can serve as ingredients for larger constructions. Table~\ref{tab:codes} evaluates GPT-6 Astra at medium effort on $(n,m)=(30,1),(36,2),(42,3),(56,1)$, complementing the larger instances in the main evaluation.

Astra matches the construction baselines at lengths 30, 36 and 42 with valid 729-word codes, and triples the reference at length 56 with a 6,561-word code. These finite constructions can be represented explicitly, while algebraic certificates (Appendix~\ref{app:trifference_certificates}) extend evaluation to larger codes whose words cannot all be listed within a response.
\input{tables/code_extension}
\appendixpage
\subsection{Claude Opus 5 on shared instances}
\label{app:opus_subset}
We also report an auxiliary comparison with the earlier Claude Opus 5 checkpoint at medium effort on 45 shared instances.
\begin{center}
\centering\footnotesize
\input{tables/opus_matched}
\captionof{table}{Comparison with Claude Opus 5 at medium effort on 45 shared instances: 15 with published frontiers and 30 with construction baselines. The overall score is 100 times the mean relative quality over this subset.}
\label{tab:opus_subset}
\end{center}
\subsection{Sensitivity to family selection}
\label{app:selection}
We examine the effect of including Kakeya as a scored family in Table~\ref{tab:selection}. This comparison uses the twelve non-code families, with and without Kakeya, on identical model responses. Spherical-code variants share one family label, as do the two Heilbronn domains; each distinct task and parameter setting retains equal weight. Kakeya affects only the mean relative quality on instances with construction baselines and the mean gap closed, leaving the published-frontier results unchanged.
\begin{center}
\centering\footnotesize
\resizebox{\linewidth}{!}{\input{tables/selection_comparison}}
\captionof{table}{Effect of including Kakeya in the aggregate for the eleven configurations with control results. $+K$ includes Kakeya and $-K$ excludes it. Relative quality is averaged over instances with construction baselines; both settings use the same responses from the twelve non-code families and equal instance weights. Published-frontier scores are identical in the two cases.}
\label{tab:selection}
\end{center}

\appendixpage
\subsection{Gap closed and progress toward the LABS target}
\label{app:gap_breakdown}
Gap closed starts at the same reference as relative quality and measures logarithmic progress toward a proven bound. Table~\ref{tab:gap_breakdown} separates the two reference groups. The combined value gives each eligible instance equal weight, so the groups contribute in proportion to their 30 and 34 instances. Invalid answers and valid answers that do not exceed the reference contribute zero.

The five LABS instances instead use the conjectured merit-factor target of 12.32. We compute progress from each construction baseline toward this target using the same logarithmic formula, and report it separately. These five values do not enter the proven-bound mean in Table~\ref{tab:headline}.
\begin{center}
\footnotesize\setlength{\tabcolsep}{4pt}
\input{tables/gap_closed_breakdown}
\captionof{table}{Mean progress from the reference. The first three numeric columns use proven mathematical bounds; the last uses the conjectured LABS target. Each mean includes zero scores.}
\label{tab:gap_breakdown}
\end{center}

%% file: tables/ladder_parameters.tex
\begin{longtable}{@{}P{.29\linewidth}P{.38\linewidth}rr@{}}
\caption{Problem-size experiments: instances and references. An asterisk marks a construction baseline.}\label{tab:ladder_parameters}\\
\toprule
Family & Parameters & Reference & Baseline \\
\midrule\endfirsthead
\toprule
Family & Parameters & Reference & Baseline \\
\midrule\endhead
\bottomrule\endfoot
cap set & d=6 & 112 & 112 \\
cap set & d=8 & 512 & 512 \\
cap set & d=10 & 2432 & 2240 \\
cap set & d=12 & 12928 & 12544 \\
spherical codes ($60^\circ$) & d=8 & 240 & 240 \\
spherical codes ($60^\circ$) & d=10 & 510 & 510 \\
spherical codes ($60^\circ$) & d=13 & 1154 & 1154 \\
spherical codes ($60^\circ$) & d=16 & 4320 & 4320 \\
max. degree 3 graphs & d=3, k=4 & 38 & 16 \\
max. degree 3 graphs & d=3, k=6 & 132 & 24 \\
max. degree 3 graphs & d=3, k=8 & 360 & 46 \\
max. degree 3 graphs & d=3, k=10 & 1250 & 94 \\
Schur & k=5 & 160 & 121 \\
Schur & k=6 & 536 & 364 \\
Schur & k=7 & 1696 & 1093 \\
Schur & k=8 & 5362 & 3280 \\
covering & v=12, k=6, t=3 & 15 & 16 \\
covering & v=16, k=6, t=3 & 38 & 48 \\
covering & v=24, k=6, t=3 & 116 & 181 \\
covering & v=30, k=6, t=3 & 225 & 366 \\
max. degree 4 graphs & d=4, k=4 & 104 & 41 \\
max. degree 4 graphs & d=4, k=5 & 364 & 61 \\
max. degree 4 graphs & d=4, k=6 & 745 & 83 \\
max. degree 4 graphs & d=4, k=7 & 1320 & 113 \\
integer AP-free sets & k=3, n=300 & 52* & 52 \\
integer AP-free sets & k=3, n=2000 & 165* & 165 \\
integer AP-free sets & k=3, n=8000 & 384* & 384 \\
integer AP-free sets & k=3, n=32000 & 1024* & 1024 \\
\end{longtable}

%% file: tables/ladder_scores.tex
\begin{longtable}{@{}P{.23\linewidth}lrrrr@{}}
\caption{Mean relative quality at each problem size. Parentheses give valid responses out of the total; means include zero scores.}\label{tab:ladder}\\
\toprule
Family & Size & \shortstack{Qwen3.8-27B\\high} & \shortstack{GPT-5.6 Luna\\high} & \shortstack{DeepSeek V4.1\\Flash low} & \shortstack{GPT-6 Astra\\high} \\
\midrule\endfirsthead
\toprule
Family & Size & \shortstack{Qwen3.8-27B\\high} & \shortstack{GPT-5.6 Luna\\high} & \shortstack{DeepSeek V4.1\\Flash low} & \shortstack{GPT-6 Astra\\high} \\
\midrule\endhead
\bottomrule\endfoot
cap set & $d=6$ & 0.00 (0/2) & 0.36 (1/2) & 0.65 (2/2) & 1.00 (2/2) \\
cap set & $d=8$ & 0.32 (1/2) & 0.78 (2/2) & 0.57 (2/2) & 0.91 (2/2) \\
cap set & $d=10$ & 0.63 (2/2) & 0.63 (2/2) & 0.60 (2/2) & 0.88 (2/2) \\
cap set & $d=12$ & 0.31 (1/2) & 0.51 (2/2) & 0.41 (2/2) & 0.97 (2/2) \\
spherical codes ($60^\circ$) & $d=8$ & 1.00 (2/2) & 0.00 (0/2) & 1.00 (2/2) & 1.00 (2/2) \\
spherical codes ($60^\circ$) & $d=10$ & 0.30 (1/2) & 0.67 (2/2) & 0.82 (2/2) & 0.98 (2/2) \\
spherical codes ($60^\circ$) & $d=13$ & 0.36 (1/2) & 0.52 (2/2) & 0.27 (2/2) & 0.46 (1/2) \\
spherical codes ($60^\circ$) & $d=16$ & 0.26 (1/2) & 0.11 (2/2) & 0.26 (1/2) & 0.53 (2/2) \\
max. degree 3 graphs & $k=4$ & 0.32 (1/2) & 0.39 (1/2) & 0.32 (1/2) & 0.89 (2/2) \\
max. degree 3 graphs & $k=6$ & 0.18 (1/2) & 0.20 (1/2) & 0.17 (2/2) & 0.95 (2/2) \\
max. degree 3 graphs & $k=8$ & 0.11 (1/2) & 0.20 (1/2) & 0.28 (2/2) & 0.66 (2/2) \\
max. degree 3 graphs & $k=10$ & 0.09 (2/2) & 0.17 (2/2) & 0.00 (0/2) & 0.30 (2/2) \\
Schur & $k=5$ & 0.00 (0/2) & 0.00 (0/2) & 0.76 (2/2) & 0.86 (2/2) \\
Schur & $k=6$ & 0.34 (1/2) & 0.00 (0/2) & 0.18 (1/2) & 0.75 (2/2) \\
Schur & $k=7$ & 0.06 (1/2) & 0.00 (0/2) & 0.00 (0/2) & 0.71 (2/2) \\
Schur & $k=8$ & 0.00 (0/2) & 0.02 (1/2) & 0.00 (0/2) & 0.74 (2/2) \\
covering & $v=12$ & 0.00 (0/2) & 0.72 (2/2) & 0.00 (0/2) & 1.00 (2/2) \\
covering & $v=16$ & 0.30 (1/2) & 0.70 (2/2) & 0.30 (1/2) & 0.86 (2/2) \\
covering & $v=24$ & 0.53 (2/2) & 0.26 (1/2) & 0.26 (1/2) & 0.94 (2/2) \\
covering & $v=30$ & 0.20 (1/2) & 0.00 (0/2) & 0.25 (1/2) & 0.84 (2/2) \\
max. degree 4 graphs & $k=4$ & -- & -- & 0.77 (1/1) & 0.84 (1/1) \\
max. degree 4 graphs & $k=5$ & -- & -- & 0.00 (0/1) & 0.53 (1/1) \\
max. degree 4 graphs & $k=6$ & -- & -- & 0.07 (1/1) & 0.98 (1/1) \\
max. degree 4 graphs & $k=7$ & -- & -- & 0.09 (1/1) & 0.15 (1/1) \\
integer AP-free sets & $n=300$ & 0.46 (1/2) & 0.46 (1/2) & 0.45 (1/2) & 1.00 (2/2) \\
integer AP-free sets & $n=2000$ & 0.00 (0/2) & 0.78 (2/2) & 0.44 (1/2) & 1.00 (2/2) \\
integer AP-free sets & $n=8000$ & 0.50 (1/2) & 1.00 (2/2) & 0.50 (1/2) & 1.04 (2/2) \\
integer AP-free sets & $n=32000$ & 0.50 (1/2) & 1.00 (2/2) & 0.50 (1/2) & 1.00 (2/2) \\
\end{longtable}

%% file: tables/code_multimodel.tex
\begin{tabular}{@{}lrrrrr@{}}
\toprule
configuration & \shortstack{Shannon\\relative quality} & valid & \shortstack{Trifference\\relative quality} & valid & words at $n=192$ \\
\midrule
Qwen3.5-4B high & 0.053 & 4/4 & 0.000 & 0/4 & 0 \\
Qwen3.5-9B high & 0.053 & 4/4 & 0.000 & 0/4 & 0 \\
Qwen3.5-27B high & 0.297 & 4/4 & 0.019 & 2/4 & 0 \\
Qwen3.8-27B high & 0.504 & 3/4 & 0.501 & 3/4 & $4.30\times10^{7}$ \\
GPT-5.6 Luna medium & 0.044 & 3/4 & 0.000 & 0/4 & 0 \\
GPT-5.6 Luna high & 0.705 & 4/4 & 0.250 & 1/4 & $1.05\times10^{10}$ \\
DeepSeek V4.1 Flash low & 0.705 & 4/4 & 0.500 & 3/4 & $1.43\times10^{7}$ \\
DeepSeek V4.1 Flash high & 0.705 & 4/4 & 0.250 & 1/4 & 0 \\
Claude Fable 5.1 medium & 0.250 & 1/4 & 1.000 & 4/4 & $1.05\times10^{10}$ \\
Claude Fable 5.1 high & 0.250 & 1/4 & 0.750 & 3/4 & $1.05\times10^{10}$ \\
Claude Opus 5.5 medium & 0.705 & 4/4 & 1.500 & 4/4 & $1.05\times10^{10}$ \\
Claude Opus 5.5 high & 0.678 & 3/4 & 1.500 & 4/4 & $1.05\times10^{10}$ \\
GPT-6 Astra medium & 0.705 & 4/4 & 1.000 & 4/4 & $1.05\times10^{10}$ \\
GPT-6 Astra high & 0.774 & 4/4 & 1.500 & 4/4 & $1.05\times10^{10}$ \\
\bottomrule
\end{tabular}

%% file: tables/code_detail_shannon.tex
\begin{tabularx}{\linewidth}{@{}Xrrrr@{}}
\toprule
Model / effort & $(7,24)$ & $(7,40)$ & $(9,48)$ & $(9,64)$ \\
\midrule
Qwen3.5-4B high & 0.145 & 0.037 & 0.023 & 0.007 \\
Qwen3.5-9B high & 0.145 & 0.037 & 0.023 & 0.007 \\
Qwen3.5-27B high & 0.145 & 0.037 & 1.000 & 0.007 \\
Qwen3.8-27B high & 0.713 & 0.304 & 0.000$^{\dagger}$ & 1.000 \\
GPT-5.6 Luna medium & 0.145 & 0.000$^{\dagger}$ & 0.023 & 0.007 \\
GPT-5.6 Luna high & 0.515 & 0.304 & 1.000 & 1.000 \\
DeepSeek V4.1 Flash low & 0.515 & 0.304 & 1.000 & 1.000 \\
DeepSeek V4.1 Flash high & 0.515 & 0.304 & 1.000 & 1.000 \\
Claude Fable 5.1 medium & 0.000$^{\dagger}$ & 0.000$^{\dagger}$ & 0.000$^{\dagger}$ & 1.000 \\
Claude Fable 5.1 high & 0.000$^{\dagger}$ & 0.000$^{\dagger}$ & 1.000 & 0.000$^{\dagger}$ \\
Claude Opus 5.5 medium & 0.515 & 0.304 & 1.000 & 1.000 \\
Claude Opus 5.5 high & 0.713 & 0.000$^{\dagger}$ & 1.000 & 1.000 \\
GPT-6 Astra medium & 0.515 & 0.304 & 1.000 & 1.000 \\
GPT-6 Astra high & 0.515 & 0.582 & 1.000 & 1.000 \\
\bottomrule
\end{tabularx}

%% file: tables/code_detail_trifference.tex
\begin{tabularx}{\linewidth}{@{}Xrrrr@{}}
\toprule
Model / effort & 64 & 96 & 144 & 192 \\
\midrule
Qwen3.5-4B high & 0.000$^{\dagger}$ & 0.000$^{\dagger}$ & 0.000$^{\dagger}$ & 0.000$^{\dagger}$ \\
Qwen3.5-9B high & 0.000$^{\dagger}$ & 0.000$^{\dagger}$ & 0.000$^{\dagger}$ & 0.000$^{\dagger}$ \\
Qwen3.5-27B high & 0.037 & 0.037 & 0.000$^{\dagger}$ & 0.000$^{\dagger}$ \\
Qwen3.8-27B high & 1.000 & 1.000 & 0.000$^{\dagger}$ & 0.004 \\
GPT-5.6 Luna medium & 0.000$^{\dagger}$ & 0.000$^{\dagger}$ & 0.000$^{\dagger}$ & 0.000$^{\dagger}$ \\
GPT-5.6 Luna high & 0.000$^{\dagger}$ & 0.000$^{\dagger}$ & 0.000$^{\dagger}$ & 1.000 \\
DeepSeek V4.1 Flash low & 1.000 & 0.000$^{\dagger}$ & 1.000 & 0.001 \\
DeepSeek V4.1 Flash high & 0.000$^{\dagger}$ & 1.000 & 0.000$^{\dagger}$ & 0.000$^{\dagger}$ \\
Claude Fable 5.1 medium & 1.000 & 1.000 & 1.000 & 1.000 \\
Claude Fable 5.1 high & 0.000$^{\dagger}$ & 1.000 & 1.000 & 1.000 \\
Claude Opus 5.5 medium & 3.000 & 1.000 & 1.000 & 1.000 \\
Claude Opus 5.5 high & 3.000 & 1.000 & 1.000 & 1.000 \\
GPT-6 Astra medium & 1.000 & 1.000 & 1.000 & 1.000 \\
GPT-6 Astra high & 3.000 & 1.000 & 1.000 & 1.000 \\
\bottomrule
\end{tabularx}

%% file: tables/code_extension.tex
\begin{center}\begin{minipage}{\linewidth}\centering\small
\begin{tabular}{@{}rrrrrc@{}}
\toprule
Length $n$ & Separation $m$ & Reference & Model & \shortstack{Relative\\quality} & Valid \\
\midrule
30 & 1 & 729 & 729 & 1.000 & yes \\
36 & 2 & 729 & 729 & 1.000 & yes \\
42 & 3 & 729 & 729 & 1.000 & yes \\
56 & 1 & 2187 & 6561 & 3.000 & yes \\
\bottomrule
\end{tabular}
\captionof{table}{GPT-6 Astra at medium effort on four smaller trifference instances. Code sizes are compared with construction baselines.}
\label{tab:codes}
\end{minipage}\end{center}

%% file: tables/opus_matched.tex
\begin{tabular}{@{}lrrrr@{}}
\toprule
& & \multicolumn{2}{c}{Mean relative quality} & \\
\cmidrule(lr){3-4}
configuration & \shortstack{Overall\\score} & \shortstack{Published\\frontiers (15)} & \shortstack{Construction\\baselines (30)} & \shortstack{Valid\\fraction} \\
\midrule
Claude Opus 5 medium & 52.41 & 0.48 & 0.55 & 0.87 \\
Qwen3.5-4B high & 7.69 & 0.08 & 0.07 & 0.38 \\
Qwen3.5-9B high & 16.96 & 0.14 & 0.18 & 0.47 \\
Qwen3.5-27B high & 17.07 & 0.17 & 0.17 & 0.53 \\
Qwen3.8-27B high & 40.98 & 0.37 & 0.43 & 0.64 \\
GPT-5.6 Luna medium & 20.36 & 0.18 & 0.21 & 0.64 \\
GPT-5.6 Luna high & 32.77 & 0.23 & 0.37 & 0.71 \\
DeepSeek V4.1 Flash low & 34.30 & 0.43 & 0.30 & 0.80 \\
DeepSeek V4.1 Flash high & 45.40 & 0.34 & 0.51 & 0.78 \\
Claude Fable 5.1 medium & 75.64 & 0.54 & 0.86 & 0.87 \\
Claude Fable 5.1 high & 36.89 & 0.54 & 0.29 & 0.44 \\
Claude Opus 5.5 medium & 60.16 & 0.58 & 0.61 & 0.87 \\
Claude Opus 5.5 high & 71.38 & 0.72 & 0.71 & 0.87 \\
GPT-6 Astra medium & 74.57 & 0.60 & 0.82 & 0.98 \\
GPT-6 Astra high & 91.48 & 0.74 & 1.00 & 1.00 \\
\bottomrule
\end{tabular}

%% file: tables/selection_comparison.tex
\begin{tabular}{@{}lrrrr@{}}
\toprule
configuration & \shortstack{Relative quality\\$+K$} & \shortstack{Relative quality\\$-K$} & \shortstack{Gap closed\\$+K$} & \shortstack{Gap closed\\$-K$} \\
\midrule
Qwen3.5-4B high & 0.088 & 0.096 & 0.000 & 0.000 \\
Qwen3.5-9B high & 0.191 & 0.202 & 0.000 & 0.000 \\
Qwen3.5-27B high & 0.184 & 0.192 & 0.000 & 0.000 \\
Qwen3.8-27B high & 0.425 & 0.418 & 0.005 & 0.004 \\
GPT-5.6 Luna medium & 0.203 & 0.207 & 0.000 & 0.000 \\
GPT-5.6 Luna high & 0.419 & 0.408 & 0.002 & 0.002 \\
DeepSeek V4.1 Flash low & 0.283 & 0.292 & 0.000 & 0.000 \\
DeepSeek V4.1 Flash high & 0.470 & 0.510 & 0.006 & 0.007 \\
Claude Fable 5.1 medium & 0.880 & 0.868 & 0.013 & 0.014 \\
GPT-6 Astra medium & 0.838 & 0.822 & 0.006 & 0.007 \\
GPT-6 Astra high & 1.015 & 1.014 & 0.017 & 0.017 \\
\bottomrule
\end{tabular}

%% file: tables/gap_closed_breakdown.tex
\begin{tabular}{@{}lrrrr@{}}
\toprule
& \multicolumn{3}{c}{Gap closed to proven bounds} & LABS target \\
\cmidrule(lr){2-4}
Configuration & All 64 & Published 30 & Construction 34 & 5 instances \\
\midrule
Qwen3.5-4B high & 0.000 & 0.000 & 0.000 & 0.000 \\
Qwen3.5-9B high & 0.000 & 0.000 & 0.000 & 0.000 \\
Qwen3.5-27B high & 0.000 & 0.000 & 0.000 & 0.000 \\
Qwen3.8-27B high & 0.004 & 0.000 & 0.007 & 0.000 \\
GPT-5.6 Luna medium & 0.000 & 0.000 & 0.000 & 0.000 \\
GPT-5.6 Luna high & 0.002 & 0.000 & 0.004 & 0.000 \\
DeepSeek V4.1 Flash low & 0.000 & 0.000 & 0.000 & 0.000 \\
DeepSeek V4.1 Flash high & 0.006 & 0.000 & 0.011 & 0.000 \\
Claude Fable 5.1 medium & 0.012 & 0.000 & 0.023 & 0.000 \\
Claude Fable 5.1 high & 0.003 & 0.000 & 0.006 & 0.000 \\
Claude Opus 5.5 medium & 0.006 & 0.000 & 0.011 & 0.000 \\
Claude Opus 5.5 high & 0.009 & 0.000 & 0.017 & 0.000 \\
GPT-6 Astra medium & 0.006 & 0.000 & 0.011 & 0.000 \\
GPT-6 Astra high & 0.016 & 0.000 & 0.030 & 0.006 \\
\midrule
GPT-6 Astra high + tools & 0.063 & 0.000 & 0.118 & 0.401 \\
GPT-5.6 Luna high + tools & 0.047 & 0.000 & 0.088 & 0.247 \\
Claude Opus 5.5 high + tools & 0.075 & 0.000 & 0.140 & 0.186 \\
\bottomrule
\end{tabular}

%% file: appendix/verification.tex
\appendixpage
\section{Verification of constructions}
\label{app:formats}
The accepted formats can express constructions that match the references for all 69 evaluated instances. Appendix~\ref{app:reference_witnesses} documents these reference constructions; the representations below include explicit objects, compact constructions and algebraic certificates.
\subsection{Verifier validation}
\label{app:verifier_validation}
The verifiers use separate code for mathematical checks and objective calculations, and share code for extracting answers and expanding compact descriptions. They agree for all model constructions that pass the primary verifier and all 69 reference constructions.

Tests include forbidden configurations, duplicates, incorrect dimensions and types, false cardinality claims and invalid certificates. Spherical-code tests exercise exact angle boundaries. For small Shannon and trifference codes, exhaustive checks compare verifier decisions with the defining constraints, and certificate tests compare the certified result with an explicitly expanded code. All deliberately invalid test cases are rejected, and the exhaustive and certificate-expansion checks agree with the verifiers. These checks and expected outcomes accompany the release.

\subsection{Compact representations}
\begin{taskbox}[title=Mathematical ingredients]
Compact descriptions specify a construction through its mathematical ingredients:
\begin{itemize}
\item \emph{Orbits}: apply coordinate permutations and sign changes to generator vectors, with duplicate vectors counted once.
\item \emph{Products}: take the Cartesian product of smaller cap sets.
\item \emph{Digit sets}: select integers by their base-$b$ digits, optionally restricting the sum of squared digits as in Behrend-type constructions.
\item \emph{Difference sets}: form a corner-free grid $\{(x,y):x-y\in D\}$ from a specified set $D$.
\item \emph{Cayley graphs}: specify cyclic-group orders and generators, including inverses for undirected edges.
\end{itemize}
Descriptions are expanded before verification, except for the algebraic code certificates below. Expansion is limited to 60{,}000 elements, with any smaller family-specific limit taking precedence. Digit enumeration allows at most 480{,}000 search nodes. Expansion and verification together have a 60-second limit; exceeding it scores zero. Prompts state the accepted representation.
\end{taskbox}
\paragraph{Exact quadratic coordinates.}
For the thirteen-dimensional kissing instance, a coordinate may be written as $a+b\sqrt3$, encoded by an integer pair $(a,b)$ with $|a|,|b|\le1000$. Both verifiers check the angle constraints exactly in $\mathbb Q(\sqrt3)$, without floating-point tolerances. This format represents the full 1,154-point published construction~\citep{zinoviev1999contact}; its JSON answer fits within 128k tokens and passes both verifiers within the 60-second limit. Submissions may contain at most 20,000 vectors.
\subsection{Code products and explicit codewords}
Shannon codes may be listed explicitly, expanded as products of up to 4,096 words, or specified by factors $C_i$ and integer powers $r_i$. The factors must satisfy $\sum_i r_i\dim(C_i)=d$. Verifying each factor establishes the full product's separation property, and its exact size is $\prod_i|C_i|^{r_i}$. Certificates may list at most 16,384 factor words in total and describe dimension at most 256. The supported alphabets are $q\in\{5,7,9\}$, with $C_5$ used for calibration. We compute a rational upper bound on $\vartheta(C_q)^d$~\citep{lovasz1979shannon} and round it down to an integer, since code sizes are integers.

The explicit-expansion format for trifference accepts up to 6,561 words of length at most 128, or a ternary generator matrix with at most eight rows. For a verified linear code, translating a triple to include zero reduces verification to pairs of codewords. Nonlinear codes require the direct triple check. Both procedures have the same 60-second limit.
\appendixpage
\subsection{Linear trifference certificates}
\label{app:trifference_certificates}
The main trifference instances use algebraic certificates to represent codes beyond the explicit-list limit. A linear-code certificate specifies a ternary generator matrix with at most 16 rows, rank $k\le12$, length $n\le1024$ and $m\in\{1,2,3\}$. The verifier computes the rank and certifies the exact size $3^k$.
\begin{constructionbox}[title=Hyperplane verification]
For a full-rank ternary generator matrix $G$, translate a triple to $(0,uG,vG)$ and put $z=u+v$, $a=u-v$. Coordinate $j$ separates the triple exactly when $(zG)_j=0$ and $(aG)_j\ne0$. The case $z=0$ requires the full code to have minimum distance at least $m$. For each nonzero projective $z$, restrict $G$ to the coordinates where $zG$ is zero. The remaining requirement is equivalent to that restriction having rank $k-1$ and minimum distance at least $m$. This extends the minimal-code characterisation of trifference~\citep{bishnoi2024trifferent,bishnoi2025generalized} into a direct check.

Verification enumerates $(3^k-1)/2$ projective directions. For $m\le3$, reduced row-echelon form allows exact distance checking using only combinations with at most $m-1$ nonzero pivot coefficients. A second verifier checks the rank and distance conditions independently.
\end{constructionbox}
\input{tables/trifference_verifier}
\begin{minipage}{\linewidth}
\subsection{Reed--Solomon concatenation}\label{app:representation_limits}
\begin{constructionbox}[title=Construction and separation guarantee]
A Reed--Solomon concatenation certificate supplies a verified inner code of length $a$ and size $q=3^r$, a monic irreducible polynomial defining $\mathbb F_q$, and outer parameters $1\le K\le N\le q$. Every triple of distinct inner codewords must be separated in at least $b$ coordinates. The outer code evaluates all polynomials of degree below $K$ at $N$ distinct field elements. Replacing its symbols by the $q$ inner words gives exactly $q^K$ words of length $aN$, with zero padding to $n$ allowed.

The separation guarantee follows from the outer code's distance. Two distinct outer polynomials agree at most $K-1$ times, so their distance is at least $D=N-K+1$. If three outer words have $t$ coordinates where all three symbols differ, their three pairwise distances sum to at most $2N+t$ and at least $3D$. Hence $t\ge N-3K+3$, and the certificate is valid when $b(N-3K+3)\ge m$. This gives an exact construction and count without expanding its words.

Both checkers verify the inner code and field irreducibility. For irreducibility, one implementation uses Frobenius/gcd tests and the other uses monic trial division. The representation supports a single concatenation layer.
\end{constructionbox}
\begin{resultbox}[title=What the certificate establishes]
\begin{tabularx}{\linewidth}{@{}XXX@{}}
Length & Number of words & Guaranteed separation\\
$aN$ & $q^K$ & $b(N-3K+3)$
\end{tabularx}
\end{resultbox}
The certificate is accepted when its guaranteed separation $b(N-3K+3)$ is at least $m$. The finite inner code and irreducible polynomial make both the cardinality and the separation claim checkable.
\end{minipage}

\paragraph{Representation limits.}
At length 64, the \texttt{bench-v1.0} formats admit at most $3^{12}=531{,}441$ words (relative quality 27), attained by Opus 5.5. Direct and explicit formats obey this cap. For concatenation, $3^r\le\lfloor2(3/2)^a\rfloor$, $N\le\min(3^r,\lfloor64/a\rfloor)$, $K\le\lfloor(N+2)/3\rfloor$ and $r\le12$ give $rK\le12$. This is a representation limit, not a known mathematical optimum; scores are not clipped at reference parity.

%% file: tables/trifference_verifier.tex
\begin{center}\begin{minipage}{\linewidth}\centering\small
\begin{tabular}{@{}lrrr@{}}
\toprule
test construction & words & primary (s) & independent (s) \\
\midrule
rank 12, $m=1$ & 531,441 & 5.61 & 8.66 \\
rank 12, $m=3$ & 531,441 & 21.63 & 37.70 \\
concatenation, $n=1024$ & $5.154\times10^{47}$ & $<0.01$ & $<0.01$ \\
\bottomrule
\end{tabular}
\captionof{table}{Measured verification times for reference constructions. Both implementations check the certificate and exact code size. The length-1024 example illustrates verification at a scale beyond the evaluated model answers.}
\label{tab:trifference_verifier}
\end{minipage}\end{center}

%% file: appendix/references.tex
\appendixpage
\section{References, bounds and targets}
\label{app:lit}
\subsection{Verified reference constructions}
\label{app:reference_witnesses}
For every one of the 69 evaluated instances, we verified a construction matching its reference, covering all 30 published frontiers and all 39 construction baselines. Each object uses an accepted answer format, fits within the 128k output limit and passes both verifiers within 60 seconds. Together, these constructions establish that a relative quality of $1$ is attainable throughout the evaluation under the allowed answer formats and verification time. All 69 objects and their verification results are included with the benchmark.
\input{tables/reference_witnesses}
\subsection{Published constructions}
Table~\ref{tab:lit} gives selected published frontiers and the proven bounds used to measure gap closed. Values follow the listed parameter order.
\begin{center}\small\setlength{\tabcolsep}{4pt}
\begin{tabularx}{\linewidth}{@{}P{.20\linewidth}YY@{}}
\toprule
Family / parameters & Reference and source & Bound and source\\
\midrule
Cap set, $d=10,11$ & 2432, 5504; Edel and Karapetyan--Karapetyan constructions~\citep{edel2004capset,karapetyan2023caps}. & 5619, 16857; proven bounds from \citet{versluis2017capset} and slicing. \\
\addlinespace
Kissing, $d=13,14,15$ & 1154, 1932, 2564; constructions~\citep{zinoviev1999contact,ganzhinov2025lines,leech1971sphere}. & 2064, 3174, 4853; proven bounds~\citep{leijenhorst2024clustered}. \\
\addlinespace
Matrix multiplication, $(4,4,5)$, $(5,5,5)$ & 61, 93; schemes from the Lille FMM catalogue. & 20, 25; proven lower bounds from tensor flattenings. \\
\addlinespace
Degree--diameter, $(4,5)$, $(7,3)$, $(3,7)$, $(4,4)$ & 364, 168, 196, 104; graphs from \citet{comellas_degdiam}. & 485, 302, 382, 161; proven Moore bounds~\citep{miller2013degdiam}. \\
\addlinespace
$\mathbb{F}_5^n$ AP-free, $n=5,6$ & 194, 649; affine charts of Edel's projective caps~\citep{edel_caps,elsholtz2020caps}. & 1296, 7497; proven bounds from the polynomial-method count~\citep{ellenberg2017capset}. \\
\addlinespace
MOLS, $n=10,14,18$ & 2, 4, 5; constructions~\citep{miller2024mols,todorov2012mols,abel2015mols}. & 8, 12, 17; proven $n-2$ or $n-1$ bounds, including \citet{lam1989order10}. \\
\addlinespace
Schur, $k=6,7,8$ & 536, 1696, 5362; partitions~\citep{fredricksen2000schur,rowley2021templates,bengone2026schur}. & 1896 at $k=6$; general factorial upper bound at $k=7,8$; proven. \\
\addlinespace
Covering, $(v,k,t)$: $(18,7,4)$, $(24,6,3)$, $(30,6,3)$ & 126, 116, 225; block lists from the La Jolla repository~\citep{ljcr}. & 111, 112, 210; proven lower bounds recorded in the repository~\citep{ljcr}. \\
\bottomrule
\end{tabularx}
\captionof{table}{Selected published frontiers and proven bounds. Values follow the listed parameter order; minimisation tasks use lower bounds.}\label{tab:lit}
\end{center}
\appendixpage
\subsection{Construction baselines and controls}
Construction baselines are verified outputs of documented construction and search procedures, providing reproducible comparison points across the evaluation. Published LABS sequences~\citep{dimitrov2022sequences}, spherical-code tables~\citep{cohn_spherical_tables} and Heilbronn configurations~\citep{stead_heilbronn_archive} provide stronger constructions at some evaluated parameters. References remain unchanged within \texttt{bench-v1.0}; exceeding a construction baseline alone does not establish a mathematical record.
\begin{center}\small\setlength{\tabcolsep}{4pt}
\begin{tabularx}{\linewidth}{@{}P{.20\linewidth}YY@{}}
\toprule
Family / parameters & Reference and source & Bound or target and source\\
\midrule
Kakeya control, $(q,n)=(13,3)$, $(7,4)$, $(11,4)$ & Expert constructions of sizes 703, 553, 2741~\citep{saraf2008kakeya}; construction baselines. & 595, 375, 2105; proven lower bounds~\citep{bukh2021kakeya}. \\
\addlinespace
LABS & Rotated Legendre construction or reference search~\citep{packebusch2016labs}. & Merit factor 12.32; conjectured target. \\
\addlinespace
Heilbronn (both domains) & Expert construction or reference search; see Appendix~\ref{app:families}. & $1/(n-2)$; trivial proven bound from triangulation. \\
\addlinespace
Linear-equation-free sets; corners & Digit constructions or reference search~\citep{behrend1946,ruzsa1993lineq}. & $n$ and $n^2$; trivial proven bounds. \\
\addlinespace
Spherical codes & Expert construction or reference search. & Cap-packing bound; proven. \\
\bottomrule
\end{tabularx}
\captionof{table}{Construction baselines and controls, with proven bounds or the conjectured LABS target.}\label{tab:construction_refs}
\end{center}
\subsection{Constructing the code references}
Shannon references optimise products of small codes: the Polak--Schrijver 367-word code in $C_7^5$, its projections, and the 81-word code $\{(i,j,2i+4j\bmod9):i,j\in\mathbb Z_9\}$ in $C_9^3$. The ten-second search modifies these small codes and recomputes their best products. The resulting references reach roughly $4.8\times10^{40}$ words and have short certificates. Recent recursive improvements~\citep{tandon2026shannon} suggest further composition methods. At $(q,d)=(7,24)$, the \texttt{bench-v1.0} baseline uses $367^4\times107$; the 108-word factor in $C_7^4$ due to Vesel and \v{Z}erovnik, listed in \citet[Table~1]{polak2019shannon}, gives the stronger product $367^4\times108$, explaining the corresponding tool-assisted improvement over this baseline.

For the shorter trifference instances, references include 729-word linear codes of lengths 26, 35 and 41 with separation multiplicities 1, 2 and 3, and a 2,187-word code of length 49. The first three augment a published generator array with selected coordinates; the last uses projective columns of weight one or two. Zero padding gives codes at the required lengths.

At lengths 64, 96, 144 and 192 with $m=1$, the reference library combines 27-word and 81-word inner codes of lengths 9 and 14, finite-geometry constructions, and a 6,561-word length-56 construction generated by GPT-6 Astra. The source constructions are included in the release. References maximise code size over the library's inner codes and outer parameters and were fixed before evaluating these lengths. This procedure does not use the search-time budget, so trifference is excluded from the timed comparison. Astra medium matches all four references through concatenation (Table~\ref{tab:trifference_scaling}). Better inner codes can improve these results. Table~\ref{tab:trifference_verifier} shows that both verifiers also handle much larger constructions within 60 seconds.
\input{tables/trifference_scaling}
\appendixpage
\subsection{Reference searches and algebraic constructions}
\label{app:search_audit}
Expert constructions can leave the baseline unchanged even when search improves. Let $r_t$ be the search objective after $t$ seconds and $z_t$ the better of that result and the expert construction. Table~\ref{tab:search_audit} compares both quantities at 10 and 600 seconds using the same expert construction. In dimension ten, for example, cap-set search improves from 1140 to 1202 points, both below the 2240-point product construction. The baseline therefore stays at 2240. For Schur with six, seven and eight colours, both search budgets return 364, 1,093 and 3,280 elements, respectively. The longer covering searches reduce the numbers of blocks from 179, 366 and 181 to 178, 361 and 178, respectively. Shannon search leaves all four code sizes unchanged. The reported gains concern the specific search procedures tested.

\paragraph{Algebraic construction baselines.}
For corners, translated ternary digit sets give AP-free difference sets $D$ and corner-free grids $\{(x,y):x-y\in D\}$. For spherical codes, candidate vectors combine selected nonzero coordinates and signs, binary sign vectors and coordinate axes; pairwise angles are checked exactly. These procedures were calibrated on 17 separate parameter settings after model responses were collected, then applied uniformly to all 20 evaluation instances.

Every reference construction passes both verifiers and is shared by all models. Calibration is separate from the timed searches, which use the same algebraic baseline at both budgets.
\begin{center}
\centering\footnotesize\setlength{\tabcolsep}{3pt}
\resizebox{\linewidth}{!}{\input{tables/search_budget_table}}
\captionof{table}{Search gains for seeds 0--2, with all four evaluated instances included for Shannon. R denotes search-resistant families; S denotes families that may benefit from search, for the procedures studied here. $\Delta_r$ compares raw 600- and 10-second search objectives; $\Delta_z$ compares the baselines after including the expert construction. Both are log improvement ratios, with direction adjusted for minimisation. Repeats are averaged within each instance before computing the means; maxima are over paired runs.}
\label{tab:search_audit}
\end{center}

\begin{center}
\centering
\includegraphics[width=\linewidth]{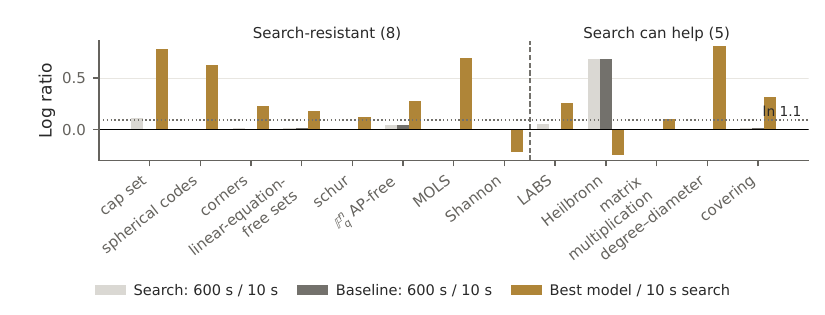}
\captionof{figure}{Effect of increasing reference search time. Light grey shows the improvement in the search result from 10 to 600 seconds. Dark grey shows the improvement after taking the better of the search result and the established construction at each budget. Ochre compares the best valid tool-free answer with the 10-second search result. For minimisation, ratios are inverted before taking logarithms; log ratios are then averaged over distinct instances. The dotted line marks a 1.1-fold improvement.}
\label{fig:search}
\end{center}

%% file: tables/reference_witnesses.tex
\begin{center}\small
\begin{tabular}{@{}lrr@{}}
\toprule
Reference group & Verified instances & Largest answer (tokens) \\
\midrule
Published frontiers & 30 / 30 & 77,528 \\
Construction baselines & 39 / 39 & 19,149 \\
\midrule
\textbf{Total} & \textbf{69 / 69} & \textbf{77,528} \\
\bottomrule
\end{tabular}
\captionof{table}{Reference constructions for all 69 instances. Each object attains relative quality $1$ and passes both verifiers within 60 seconds. Token counts use the larger of \texttt{cl100k\_base} and \texttt{o200k\_base} for the serialised answer.}\label{tab:reference_witnesses}
\end{center}

%% file: tables/trifference_scaling.tex
\begin{center}\begin{minipage}{\linewidth}\centering\small
\begin{tabular}{@{}lrrrrr@{}}
\toprule
instance & length & reference words & verified words & \shortstack{Relative\\quality} & generation (s) \\
\midrule
T1 & 64 & 19,683 & 19,683 & 1.000 & 52.3 \\
T2 & 96 & 531,441 & 531,441 & 1.000 & 53.8 \\
T3 & 144 & 387,420,489 & 387,420,489 & 1.000 & 37.0 \\
T4 & 192 & 10,460,353,203 & 10,460,353,203 & 1.000 & 63.8 \\
\bottomrule
\end{tabular}
\captionof{table}{GPT-6 Astra at medium effort on four trifference lengths. The evaluator checks each submitted construction and computes its exact size. Ratios use the construction baselines; invalid or budget-exhausted responses score zero.}
\label{tab:trifference_scaling}
\end{minipage}\end{center}

%% file: tables/search_budget_table.tex
\begin{tabular}{@{}llrrrrrr@{}}
\toprule
family & class & runs & instances & mean $\Delta_r$ & max $\Delta_r$ & mean $\Delta_z$ & max $\Delta_z$ \\
\midrule
cap set & R & 3 & 2 & 0.108 & 0.163 & 0.000 & 0.000 \\
spherical codes & R & 6 & 5 & 0.000 & 0.000 & 0.000 & 0.000 \\
corners & R & 3 & 3 & 0.019 & 0.024 & 0.000 & 0.000 \\
linear-equation-free sets & R & 3 & 3 & 0.016 & 0.019 & 0.016 & 0.019 \\
Schur & R & 3 & 3 & 0.000 & 0.000 & 0.000 & 0.000 \\
$\mathbb{F}_q^n$ AP-free & R & 3 & 2 & 0.041 & 0.083 & 0.041 & 0.083 \\
MOLS & R & 3 & 3 & 0.000 & 0.000 & 0.000 & 0.000 \\
Shannon & R & 4 & 4 & 0.000 & 0.000 & 0.000 & 0.000 \\
\midrule
LABS & S & 3 & 3 & 0.058 & 0.086 & 0.000 & 0.000 \\
Heilbronn & S & 6 & 6 & 0.687 & 0.914 & 0.687 & 0.914 \\
matrix multiplication & S & 3 & 1 & 0.000 & 0.000 & 0.000 & 0.000 \\
degree--diameter & S & 3 & 2 & 0.000 & 0.000 & 0.000 & 0.000 \\
covering & S & 3 & 3 & 0.012 & 0.017 & 0.012 & 0.017 \\
\bottomrule
\end{tabular}

%% file: appendix/parameters.tex
\appendixpage
\section{Instance parameters and resource limits}
\label{app:tiers}
We use three parameter ranges: A1 contains small familiar instances, while A2 and A3 increase size. Family selection for A3 screened the first sampled setting for a ratio of at least 1.4 between the implemented baseline $z$ and a bound or target $b$: $b/z$ for maximisation and $z/b$ for minimisation. This screening identifies families with room for improvement; the remaining instances vary the parameters within those families. Answers must also fit the representation limits and be verifiable within 60 seconds. Kakeya is evaluated separately as a control. For example, the Kakeya setting $(q,n)=(13,3)$ has a proven lower bound of 595~\citep{bukh2021kakeya} and a construction of size 703, leaving a ratio of only 1.18.

The main evaluation contains 69 instances: 61 in the twelve non-code families, four Shannon instances and four trifference instances. Each configuration contributes one response per instance under the protocol in Appendix~\ref{app:runner}. The released manifest specifies the selected instances. Table~\ref{tab:tiers} lists the instances, and Table~\ref{tab:triangle_domains} specifies the triangular domains. Kakeya provides a separate control for the family-selection comparison. Smaller-instance parameters appear in Table~\ref{tab:aux_parameters}.
\small\setlength{\tabcolsep}{4pt}
\input{tables/formal_parameters}
\normalsize
\subsection{Triangular domains}
\small\input{tables/triangle_domains}\normalsize
Each triangle is specified by its three integer-coordinate vertices. The objective is the minimum triangle area divided by the area of the containing domain.
\appendixpage
\subsection{Smaller-instance parameter draws}
\small\input{tables/auxiliary_parameters}\normalsize

%% file: tables/formal_parameters.tex
\begin{longtable}{@{}P{.30\linewidth}P{.67\linewidth}@{}}
\caption{The 69 instances in the main evaluation. Each configuration contributes one response per instance.}\label{tab:tiers}\\
\toprule
Family & Parameters \\
\midrule\endfirsthead
\toprule
Family & Parameters \\
\midrule\endhead
\bottomrule\endfoot
Cap sets & $d=10$ \\
Cap sets & $d=11$ \\
Cap sets & $d=9$ \\
\addlinespace[4pt]
Corners & $n=138$ \\
Corners & $n=183$ \\
Corners & $n=150$ \\
Corners & $n=162$ \\
Corners & $n=185$ \\
\addlinespace[4pt]
Schur & $k=7$ \\
Schur & $k=6$ \\
Schur & $k=8$ \\
\addlinespace[4pt]
$\mathbb F_q^n$ AP-free & $n=6,\ q=5$ \\
$\mathbb F_q^n$ AP-free & $n=5,\ q=5$ \\
$\mathbb F_q^n$ AP-free & $n=4,\ q=7$ \\
\addlinespace[4pt]
Spherical ($60^\circ$) & $d=14$ \\
Spherical ($60^\circ$) & $d=15$ \\
Spherical (variable) & $c=24/100,\ d=12$ \\
Spherical (variable) & $c=39/100,\ d=12$ \\
Spherical (variable) & $c=36/100,\ d=14$ \\
Spherical (variable) & $c=39/100,\ d=14$ \\
Spherical (variable) & $c=36/100,\ d=12$ \\
Spherical ($60^\circ$) & $d=13$ \\
\addlinespace[4pt]
LABS & $N=280$ \\
LABS & $N=375$ \\
LABS & $N=357$ \\
LABS & $N=281$ \\
LABS & $N=240$ \\
\addlinespace[4pt]
Heilbronn (square) & $n=46$ \\
Heilbronn (square) & $n=26$ \\
Heilbronn (square) & $n=40$ \\
Heilbronn (square) & $n=50$ \\
Heilbronn (square) & $n=48$ \\
Heilbronn (triangle) & $n=32$, $T_{1}$ \\
Heilbronn (triangle) & $n=34$, $T_{2}$ \\
Heilbronn (triangle) & $n=31$, $T_{3}$ \\
Heilbronn (triangle) & $n=40$, $T_{4}$ \\
Heilbronn (triangle) & $n=26$, $T_{5}$ \\
\addlinespace[4pt]
Degree--diameter & $d=4,\ k=5$ \\
Degree--diameter & $d=7,\ k=3$ \\
Degree--diameter & $d=4,\ k=4$ \\
Degree--diameter & $d=3,\ k=7$ \\
Degree--diameter & $d=6,\ k=3$ \\
Degree--diameter & $d=5,\ k=4$ \\
Degree--diameter & $d=3,\ k=6$ \\
\addlinespace[4pt]
Linear-equation-free & $a=3,\ b=4,\ n=19044$ \\
Linear-equation-free & $a=2,\ b=5,\ n=18883$ \\
Linear-equation-free & $a=2,\ b=3,\ n=17965$ \\
Linear-equation-free & $a=3,\ b=4,\ n=13312$ \\
Linear-equation-free & $a=3,\ b=4,\ n=6248$ \\
\addlinespace[4pt]
Covering designs & $k=7,\ t=4,\ v=18$ \\
Covering designs & $k=6,\ t=3,\ v=30$ \\
Covering designs & $k=6,\ t=3,\ v=24$ \\
\addlinespace[4pt]
Matrix multiplication & $m=4,\ n=4,\ p=5$ \\
Matrix multiplication & $m=5,\ n=5,\ p=5$ \\
Matrix multiplication & $m=5,\ n=4,\ p=5$ \\
\addlinespace[4pt]
MOLS & $n=14$ \\
MOLS & $n=10$ \\
MOLS & $n=18$ \\
MOLS & $n=15$ \\
MOLS & $n=12$ \\
MOLS & $n=20$ \\
\addlinespace[4pt]
Shannon & $d=24,\ q=7$ \\
Shannon & $d=40,\ q=7$ \\
Shannon & $d=48,\ q=9$ \\
Shannon & $d=64,\ q=9$ \\
\addlinespace[4pt]
Trifference & $m=1,\ n=64$ \\
Trifference & $m=1,\ n=96$ \\
Trifference & $m=1,\ n=144$ \\
Trifference & $m=1,\ n=192$ \\
\addlinespace[4pt]
\end{longtable}

%% file: tables/triangle_domains.tex
\begin{longtable}{@{}lP{.27\linewidth}P{.27\linewidth}P{.27\linewidth}@{}}
\caption{Triangular domains used in the main evaluation. Coordinates are integers.}\label{tab:triangle_domains}\\
\toprule
Domain & Vertex 1 & Vertex 2 & Vertex 3 \\
\midrule\endfirsthead
\toprule
Domain & Vertex 1 & Vertex 2 & Vertex 3 \\
\midrule\endhead
\bottomrule\endfoot
$T_{1}$ & $(9026,8197)$ & $(1769,9178)$ & $(7298,1276)$ \\
$T_{2}$ & $(578,9933)$ & $(8598,7337)$ & $(2075,588)$ \\
$T_{3}$ & $(411,7999)$ & $(9004,9890)$ & $(3031,713)$ \\
$T_{4}$ & $(7922,8793)$ & $(9917,1548)$ & $(1510,3942)$ \\
$T_{5}$ & $(526,8526)$ & $(8696,8279)$ & $(8084,1688)$ \\
\end{longtable}

%% file: tables/auxiliary_parameters.tex
\begin{longtable}{@{}P{.30\linewidth}lP{.56\linewidth}@{}}
\caption{Three representative smaller-instance draws per tier, in seed order 0, 1, 2. Trifference uses the lengths listed in the main evaluation table.}\label{tab:aux_parameters}\\
\toprule
Family & Tier & Representative parameters \\
\midrule\endfirsthead
\toprule
Family & Tier & Representative parameters \\
\midrule\endhead
\bottomrule\endfoot
Cap sets & A1 & $d=3$; $d=3$; $d=4$ \\
 & A2 & $d=8$; $d=7$; $d=7$ \\
\addlinespace[4pt]
Corner-free sets & A1 & $n=23$; $n=15$; $n=22$ \\
 & A2 & $n=66$; $n=50$; $n=79$ \\
\addlinespace[4pt]
Schur colourings & A1 & $k=4$; $k=4$; $k=3$ \\
 & A2 & $k=5$; $k=5$; $k=6$ \\
\addlinespace[4pt]
Progression-free sets over finite fields & A1 & $q=5,\ n=4$; $q=5,\ n=4$; $q=5,\ n=4$ \\
 & A2 & $q=7,\ n=3$; $q=7,\ n=4$; $q=7,\ n=4$ \\
\addlinespace[4pt]
Spherical codes (60$^\circ$) & A1 & $d=7$; $d=7$; $d=7$ \\
 & A2 & $d=11$; $d=12$; $d=9$ \\
\addlinespace[4pt]
Spherical codes (variable angle) & A1 & $d=6,\ c=37/100$; $d=6,\ c=35/100$; $d=6,\ c=48/100$ \\
 & A2 & $d=10,\ c=47/100$; $d=10,\ c=45/100$; $d=9,\ c=20/100$ \\
\addlinespace[4pt]
Low-autocorrelation binary sequences & A1 & $N=19$; $N=25$; $N=29$ \\
 & A2 & $N=68$; $N=74$; $N=91$ \\
\addlinespace[4pt]
Heilbronn triangles (square) & A1 & $n=7$; $n=6$; $n=5$ \\
 & A2 & $n=18$; $n=17$; $n=18$ \\
\addlinespace[4pt]
Heilbronn triangles (triangle) & A1 & $n=9$; $n=12$; $n=11$ \\
 & A2 & $n=16$; $n=14$; $n=12$ \\
\addlinespace[4pt]
Degree--diameter graphs & A1 & $d=4,\ k=3$; $d=3,\ k=3$; $d=3,\ k=3$ \\
 & A2 & $d=10,\ k=2$; $d=5,\ k=3$; $d=3,\ k=5$ \\
\addlinespace[4pt]
Linear-equation-free sets & A1 & $a=3,\ b=4,\ n=129$; $a=1,\ b=2,\ n=101$; $a=2,\ b=3,\ n=155$ \\
 & A2 & $a=3,\ b=5,\ n=1394$; $a=1,\ b=3,\ n=1445$; $a=3,\ b=4,\ n=1439$ \\
\addlinespace[4pt]
Covering designs & A1 & $v=12,\ k=6,\ t=3$; $v=13,\ k=4,\ t=2$; $v=10,\ k=5,\ t=3$ \\
 & A2 & $v=16,\ k=6,\ t=3$; $v=15,\ k=7,\ t=4$; $v=16,\ k=6,\ t=3$ \\
\addlinespace[4pt]
Matrix multiplication & A1 & $n=2,\ m=2,\ p=4$; $n=3,\ m=3,\ p=3$; $n=2,\ m=2,\ p=4$ \\
 & A2 & $n=3,\ m=3,\ p=5$; $n=3,\ m=3,\ p=4$; $n=4,\ m=4,\ p=5$ \\
\addlinespace[4pt]
Mutually orthogonal Latin squares & A1 & $n=12$; $n=10$; $n=10$ \\
 & A2 & $n=15$; $n=14$; $n=14$ \\
\addlinespace[4pt]
Shannon codes & A1 & $q=7,\ d=3$; $q=9,\ d=3$; $q=7,\ d=4$ \\
 & A2 & $q=9,\ d=12$; $q=7,\ d=8$; $q=7,\ d=12$ \\
\addlinespace[4pt]
\end{longtable}

%% file: appendix/settings.tex
\appendixpage
\section{Evaluation settings and example prompt}
\label{app:runner}
\subsection{Inference settings}
The fourteen tool-free configurations follow Section~\ref{sec:protocol}. For each instance, we retain the first completed response or generation-budget stop. Completed responses are verified; invalid answers and generation-budget exhaustion score zero and are not retried. Network and service failures are unscored, and the evaluation may resume or be resubmitted after recovery. Bounded transport reconnection is permitted, so an evaluation can involve more than one generation request. Usage records include all reported tokens, with unreported usage on interrupted requests marked as unknown. Code-family, thirteen-dimensional kissing and Claude Fable 5.1 high and Opus 5.5 evaluations use a two-hour generation deadline, followed by a separate 60-second verification limit. Appendix~\ref{app:tools} describes the tool-assisted protocol.

All configurations receive identical mathematical prompts and format instructions. API models receive a mathematical-construction system message; CLI models retain their native system instructions. No follow-up prompts are used to extend completed answers. Tool-free CLI runs disable tools in their configuration: Claude Code uses \verb|--tools ""| with a strict MCP configuration, while Codex disables execution, web and MCP tools.
\paragraph{Claude Fable 5.1.} Claude Code 2.1.269 at medium and high effort, using independent sessions and the model identifier \texttt{claude-fable-5-1}. Medium-effort responses were collected on 12--17 September 2026 and high-effort responses on 20 September 2026.
\paragraph{Claude Opus 5.5.} Claude Code 2.1.280 at medium and high effort, with model identifier \texttt{claude-opus-5-5}, collected on 22--23 September 2026. Each configuration contains 69 fresh responses, with the same prompts, output budget and verification rules as the other tool-free evaluations.
\paragraph{GPT-6 Astra and GPT-5.6 Luna.} Codex CLI 0.154.0 at the reasoning-effort levels indicated in the results.
\paragraph{DeepSeek V4.1 Flash.} Chat-completions API with reasoning enabled at low or high effort.
\paragraph{Qwen models.} vLLM 0.29.0 at high reasoning effort. The 27B checkpoints use FP8 weights; serving settings are supplied with the release.
\paragraph{Reference searches.} Each search uses a fixed random seed and one process, with iterated greedy construction or local optimisation as appropriate to the family. We report results at 10 and 600 seconds.
\subsection{Example prompt}
\label{app:prompt}
Prompts specify the problem, parameters and supported answer format. The following prompt was used for the ten-dimensional cap-set instance.
\begin{taskbox}[title=Model prompt]
\input{tables/example_prompt}
\end{taskbox}
In the tool-free track, each prompt also states that tools are unavailable and requests a single JSON answer.

%% file: tables/example_prompt.tex
\begin{quote}\small\ttfamily\raggedright Problem (Cap-Set, d=10). Find a subset S of F\_3\^{}10 (vectors of length 10 with entries in \{0,1,2\}) of maximum possible size containing no three distinct vectors x, y, z with x + y + z = 0 (mod 3) coordinate-wise -- a cap set. Output: a JSON list of the vectors, each written as a string of 10 digits (at most 20000 vectors are accepted).\par Output format: a JSON object \{"answer": X\} where X is a JSON array of strings of digits 0/1/2, all of the stated length.\par An answer that violates the stated property, is malformed, contains duplicates, or is out of range scores zero. Among valid answers, a better objective scores higher. Give one answer.\par Compact form allowed: \{"product": [S1, S2, ...]\} where S\_i is a list of digit strings in dimension d\_i with d\_1 + ... = d; the set is all concatenations.\end{quote}

%% file: appendix/tools.tex
\appendixpage
\section{Code and web access}
\label{app:tools}
\readingnote{Code and web access improve all three models' overall scores. Claude Opus 5.5 achieves the highest overall score, while GPT-6 Astra comes closest to the published frontiers. Family and instance comparisons distinguish these patterns.}

\subsection{Protocol}
We evaluated GPT-6 Astra and GPT-5.6 Luna at high effort through Codex CLI 0.154.0 on 18 September 2026, with code execution and live web search. Each instance has an independent environment with four CPU threads, 16\,GiB of memory and a two-hour wall-clock limit, with no limit on total output tokens. An evaluation may contain multiple model and tool interactions. Models can inspect constructions and stop when ready, saving their submission as a JSON file of at most 32\,MiB. When the evaluation ends or reaches a resource limit, we score the saved construction if it was written within the time limit and passes verification. Missing, late or invalid submissions receive zero. Both verifiers and their 60-second verification limit are the same as in the tool-free track.

Claude Opus 5.5 was evaluated at high effort through Claude Code 2.1.280 on 22--23 September 2026. It uses native web search and an MCP shell connected to the same CPU environment and resource limits. The native system instructions and tool interfaces therefore differ between Claude Code and Codex. Claude's per-response output limit is 128k; there is no cumulative limit across the trajectory. At an output or resource stop, the saved construction is scored under the same rule.

Mathematical prompts, parameters and formats match the tool-free evaluation; we replace the no-tools instruction with: ``Save your final answer as a single JSON object to /workspace/answer.json.'' Prompts give no reference values, strategies or budget reminders, and environments contain no benchmark library, reference objects or prior answers. Subagents are disabled and bounded reconnection is enabled. Network and service failures are unscored. We retain the first result that can be scored and do not repeat an evaluation to improve its answer.

Web access permits retrieval and adaptation of published constructions. Benchmark materials were available online during the Opus 5.5 evaluation, but not during the Astra and Luna evaluations. An audit of Opus 5.5's recorded interactions found no evidence of direct access to these materials.

\subsection{Construction quality}
GPT-6 Astra returns \ToolValid{} valid constructions, improving \ToolImproved{} and tying \ToolTied{} tool-free outcomes. GPT-5.6 Luna returns \LunaToolValid{} valid constructions, with \LunaToolImproved{} improvements, \LunaToolTied{} ties and \LunaToolDeclined{} declines. Luna's three zeroes comprise two invalid submissions and one evaluation that exceeded the memory limit before producing a submission. Both verifiers agree on every scored outcome.

Opus 5.5 returns \OpusToolValid{} valid constructions, with \OpusToolImproved{} improvements and \OpusToolTied{} ties against its high-effort tool-free answers. Its final covering evaluation reaches the two-hour limit with a valid 225-block construction saved, matching the reference and receiving full relative quality. Its mean relative quality is \OpusToolPublishedHFR{} on published frontiers and \OpusToolConstructionRatio{} on construction baselines, matching \OpusToolPublishedMatched{} published references and exceeding \OpusToolConstructionAbove{} construction baselines.

Mean relative quality on instances with published frontiers reaches \ToolPublishedHFR{} for Astra and \LunaToolPublishedHFR{} for Luna, matching \ToolPublishedMatched{} and \LunaToolPublishedMatched{} references without exceeding any. On the 39 construction baselines, Astra improves \ToolConstructionAbove{} and Luna \LunaToolConstructionAbove{}, reaching mean ratios of \ToolConstructionRatio{} and \LunaToolConstructionRatio{}. Luna exceeds Astra on \LunaToolBeatsAstra{} Heilbronn instances, with a higher mean across the ten instances in that family. These comparisons include web retrieval, code execution and the different generation budgets of the two tracks.

\begin{center}\small\setlength{\tabcolsep}{3pt}
\input{tables/tool_families}
\captionof{table}{GPT-6 Astra, GPT-5.6 Luna and Claude Opus 5.5 at high effort, with and without tools on the 69 instances. Family means average relative quality over instances, including zeroes.}
\label{tab:tool_families}
\end{center}

For the length-64 trifference example in Section~\ref{sec:representative_tasks}, tools increase Astra's code size from 59,049 to \ToolTriWords{} codewords, raising relative quality from 3 to \ToolTriRatio{}. Opus 5.5 constructs \OpusToolTriWords{} codewords, with relative quality \OpusToolTriRatio{} against the same construction baseline. This reaches the maximum size supported by the \texttt{bench-v1.0} formats at length 64, rather than a known mathematical optimum (Appendix~\ref{app:representation_limits}). This instance and the ten Heilbronn instances account for more than its net score advantage over Astra; the other 58 instances slightly favour Astra. Opus remains first when any single family is excluded. Luna matches the construction baselines at lengths 64, 96 and 192, but its length-144 certificate is invalid. All submitted objects are released.

\subsection{Token usage}
Table~\ref{tab:tool_usage} sums reported token usage across all model turns in each model's 69 evaluations. Reasoning tokens are included in output totals, and cached tokens are included in input totals. For Claude Code, we also include model usage reported by its native web-search tool. The final deadline-stopped Opus evaluation lacks that aggregate, so its retained streamed usage and the resulting total are lower bounds, marked with a star. Figure~\ref{fig:score_tokens} describes model generation, including reported search-related model usage; it does not measure external CPU computation or search-service work.
\begin{center}\small
\input{tables/tool_usage}
\captionof{table}{Reported token usage at high effort with code and web access. Stars indicate totals with some unreported usage; actual consumption may be higher.}
\label{tab:tool_usage}
\end{center}

\clearpage
\subsection{Individual instances}
Parameters and reference types are listed in Appendix~\ref{app:tiers}; the release includes every submitted object and its verification result.
\small\setlength{\tabcolsep}{2pt}
\input{tables/tool_instances}
\normalsize

%% file: tables/tool_families.tex
\begin{tabularx}{\linewidth}{@{}Xrrrrrrr@{}}
\toprule
& & \multicolumn{2}{c}{Astra high} & \multicolumn{2}{c}{Luna high} & \multicolumn{2}{c}{Opus 5.5 high} \\
\cmidrule(lr){3-4}\cmidrule(lr){5-6}\cmidrule(lr){7-8}
Family & $n$ & No tools & Tools & No tools & Tools & No tools & Tools \\
\midrule
cap set & 3 & 0.923 & 0.984 & 0.464 & 0.946 & 0.877 & 0.955 \\
LABS & 5 & 0.985 & 1.374 & 0.000 & 1.189 & 0.379 & 1.166 \\
Heilbronn & 10 & 0.684 & 2.138 & 0.196 & 2.270 & 0.456 & 2.936 \\
spherical codes & 8 & 1.313 & 1.642 & 0.284 & 1.420 & 0.927 & 1.784 \\
corners & 5 & 1.026 & 1.109 & 0.961 & 1.078 & 1.009 & 1.117 \\
matrix multiplication & 3 & 0.895 & 1.000 & 0.000 & 0.886 & 0.595 & 1.000 \\
degree--diameter & 7 & 0.752 & 1.000 & 0.257 & 1.000 & 0.476 & 0.992 \\
linear-equation-free & 5 & 0.974 & 1.477 & 0.564 & 0.597 & 0.971 & 1.387 \\
$\mathbb F_q^n$ AP-free & 3 & 1.062 & 1.168 & 0.691 & 1.052 & 0.902 & 1.231 \\
MOLS & 6 & 0.550 & 1.000 & 0.433 & 0.958 & 0.592 & 0.958 \\
Schur & 3 & 0.731 & 1.000 & 0.058 & 1.000 & 0.246 & 1.000 \\
covering & 3 & 0.883 & 1.000 & 0.111 & 0.861 & 0.312 & 1.000 \\
Shannon & 4 & 0.774 & 1.002 & 0.705 & 0.998 & 0.678 & 1.002 \\
trifference & 4 & 1.500 & 3.000 & 0.250 & 0.750 & 1.500 & 7.500 \\
\bottomrule
\end{tabularx}

%% file: tables/tool_usage.tex
\begin{tabular}{@{}lrrr@{}}
\toprule
Measure & Astra high & Luna high & Opus 5.5 high \\
\midrule
Input tokens & 34{,}575{,}392 & 98{,}019{,}732 & 157{,}011{,}011$^{*}$ \\
Cached input tokens (included above) & 31{,}453{,}056 & 91{,}690{,}496 & 149{,}741{,}729$^{*}$ \\
Output tokens & 445{,}123 & 866{,}731 & 1{,}702{,}589$^{*}$ \\
Reasoning tokens (included above) & 181{,}416 & 425{,}435 & 668{,}352$^{*}$ \\
\bottomrule
\end{tabular}

%% file: tables/tool_instances.tex
\begin{longtable}{@{}P{.21\linewidth}P{.21\linewidth}rrrrrrr@{}}
\caption{Per-instance relative quality for GPT-6 Astra, GPT-5.6 Luna and Claude Opus 5.5 at high effort. P: published frontier; C: construction baseline. A dagger marks an evaluation assigned zero under the protocol in Section~\ref{sec:protocol}.}\label{tab:tool_instances}\\
\toprule & & & \multicolumn{2}{c}{Astra high} & \multicolumn{2}{c}{Luna high} & \multicolumn{2}{c}{Opus 5.5 high} \\ \cmidrule(lr){4-5}\cmidrule(lr){6-7}\cmidrule(lr){8-9} Family & Parameters & Ref. & No tools & Tools & No tools & Tools & No tools & Tools \\ \midrule \endfirsthead
\toprule & & & \multicolumn{2}{c}{Astra high} & \multicolumn{2}{c}{Luna high} & \multicolumn{2}{c}{Opus 5.5 high} \\ \cmidrule(lr){4-5}\cmidrule(lr){6-7}\cmidrule(lr){8-9} Family & Parameters & Ref. & No tools & Tools & No tools & Tools & No tools & Tools \\ \midrule \endhead
\bottomrule \endfoot
cap set & $d=9$ & P & 0.932 & 1.000 & 0.739 & 1.000 & 0.932 & 0.946 \\
cap set & $d=10$ & P & 0.921 & 1.000 & 0.000$^{\dagger}$ & 0.921 & 0.785 & 0.921 \\
cap set & $d=11$ & P & 0.916 & 0.951 & 0.654 & 0.916 & 0.916 & 0.997 \\
\addlinespace[2pt]
LABS & $N=240$ & C & 0.931 & 1.568 & 0.000$^{\dagger}$ & 1.306 & 0.931 & 1.067 \\
LABS & $N=280$ & C & 0.985 & 1.507 & 0.000$^{\dagger}$ & 1.253 & 0.000$^{\dagger}$ & 1.123 \\
LABS & $N=281$ & C & 0.989 & 1.517 & 0.000$^{\dagger}$ & 1.328 & 0.000$^{\dagger}$ & 1.517 \\
LABS & $N=357$ & C & 0.996 & 1.076 & 0.000$^{\dagger}$ & 0.877 & 0.966 & 1.068 \\
LABS & $N=375$ & C & 1.022 & 1.201 & 0.000$^{\dagger}$ & 1.183 & 0.000$^{\dagger}$ & 1.054 \\
\addlinespace[2pt]
Heilbronn (square) & $n=26$ & C & 0.774 & 2.251 & 0.320 & 2.413 & 0.272 & 2.887 \\
Heilbronn (square) & $n=40$ & C & 0.837 & 3.052 & 0.096 & 1.801 & 0.463 & 1.793 \\
Heilbronn (square) & $n=46$ & C & 1.193 & 3.018 & 0.089 & 3.232 & 0.497 & 2.066 \\
Heilbronn (square) & $n=48$ & C & 1.173 & 1.952 & 0.130 & 0.207 & 0.491 & 4.470 \\
Heilbronn (square) & $n=50$ & C & 0.793 & 2.930 & 0.238 & 4.572 & 0.685 & 5.950 \\
Heilbronn (triangle) & $n=26$, $T_{5}$ & C & 0.224 & 1.234 & 0.149 & 1.159 & 0.120 & 1.252 \\
Heilbronn (triangle) & $n=31$, $T_{3}$ & C & 0.468 & 1.903 & 0.115 & 2.066 & 0.454 & 1.630 \\
Heilbronn (triangle) & $n=32$, $T_{1}$ & C & 0.322 & 1.623 & 0.180 & 2.137 & 0.456 & 1.993 \\
Heilbronn (triangle) & $n=34$, $T_{2}$ & C & 0.425 & 1.707 & 0.250 & 2.246 & 0.535 & 2.847 \\
Heilbronn (triangle) & $n=40$, $T_{4}$ & C & 0.635 & 1.711 & 0.398 & 2.864 & 0.590 & 4.475 \\
\addlinespace[2pt]
spherical codes & $d=13$ & P & 0.924 & 1.000 & 0.270 & 1.000 & 0.827 & 1.000 \\
spherical codes & $d=14$ & P & 0.652 & 1.000 & 0.188 & 1.000 & 0.768 & 1.000 \\
spherical codes & $d=15$ & P & 0.667 & 1.000 & 0.164 & 1.000 & 0.913 & 1.000 \\
spherical codes & $c=24/100,\ d=12$ & C & 1.964 & 2.786 & 0.857 & 2.750 & 0.000$^{\dagger}$ & 2.571 \\
spherical codes & $c=36/100,\ d=12$ & C & 1.105 & 1.276 & 0.368 & 1.105 & 1.105 & 1.230 \\
spherical codes & $c=39/100,\ d=12$ & C & 1.184 & 1.533 & 0.421 & 1.000 & 1.105 & 1.520 \\
spherical codes & $c=36/100,\ d=14$ & C & 1.806 & 2.113 & 0.000$^{\dagger}$ & 1.806 & 0.000$^{\dagger}$ & 2.766 \\
spherical codes & $c=39/100,\ d=14$ & C & 2.197 & 2.432 & 0.000$^{\dagger}$ & 1.697 & 2.697 & 3.182 \\
\addlinespace[2pt]
corners & $n=138$ & C & 1.058 & 1.078 & 1.000 & 1.058 & 1.000 & 1.078 \\
corners & $n=150$ & C & 1.019 & 1.083 & 1.000 & 1.053 & 1.019 & 1.097 \\
corners & $n=162$ & C & 1.052 & 1.117 & 1.000 & 1.078 & 1.019 & 1.121 \\
corners & $n=183$ & C & 1.000 & 1.138 & 0.838 & 1.113 & 1.008 & 1.145 \\
corners & $n=185$ & C & 1.000 & 1.127 & 0.966 & 1.090 & 1.000 & 1.142 \\
\addlinespace[2pt]
matrix multiplication & $m=4,\ n=4,\ p=5$ & P & 0.938 & 1.000 & 0.000$^{\dagger}$ & 0.762 & 0.938 & 1.000 \\
matrix multiplication & $m=5,\ n=4,\ p=5$ & P & 0.894 & 1.000 & 0.000$^{\dagger}$ & 0.894 & 0.000$^{\dagger}$ & 1.000 \\
matrix multiplication & $m=5,\ n=5,\ p=5$ & P & 0.853 & 1.000 & 0.000$^{\dagger}$ & 1.000 & 0.845 & 1.000 \\
\addlinespace[2pt]
degree--diameter & $d=3,\ k=6$ & P & 0.955 & 1.000 & 0.333 & 1.000 & 0.485 & 1.000 \\
degree--diameter & $d=3,\ k=7$ & P & 0.857 & 1.000 & 0.622 & 1.000 & 0.551 & 1.000 \\
degree--diameter & $d=4,\ k=4$ & P & 0.769 & 1.000 & 0.000$^{\dagger}$ & 1.000 & 0.548 & 0.952 \\
degree--diameter & $d=4,\ k=5$ & P & 0.558 & 1.000 & 0.000$^{\dagger}$ & 1.000 & 0.558 & 1.000 \\
degree--diameter & $d=5,\ k=4$ & P & 0.802 & 1.000 & 0.000$^{\dagger}$ & 1.000 & 0.302 & 0.991 \\
degree--diameter & $d=6,\ k=3$ & P & 0.730 & 1.000 & 0.559 & 1.000 & 0.495 & 1.000 \\
degree--diameter & $d=7,\ k=3$ & P & 0.595 & 1.000 & 0.286 & 1.000 & 0.393 & 1.000 \\
\addlinespace[2pt]
linear-equation-free & $a=3,\ b=4,\ n=6248$ & C & 1.032 & 1.338 & 0.135 & 1.020 & 0.000$^{\dagger}$ & 1.433 \\
linear-equation-free & $a=3,\ b=4,\ n=13312$ & C & 0.766 & 1.445 & 0.822 & 0.000$^{\dagger}$ & 1.199 & 1.248 \\
linear-equation-free & $a=2,\ b=3,\ n=17965$ & C & 0.717 & 1.421 & 0.478 & 1.036 & 0.964 & 1.423 \\
linear-equation-free & $a=2,\ b=5,\ n=18883$ & C & 1.576 & 1.576 & 0.841 & 0.931 & 1.576 & 1.603 \\
linear-equation-free & $a=3,\ b=4,\ n=19044$ & C & 0.778 & 1.603 & 0.544 & 0.000$^{\dagger}$ & 1.117 & 1.230 \\
\addlinespace[2pt]
$\mathbb F_q^n$ AP-free & $n=4,\ q=7$ & C & 1.417 & 1.505 & 1.427 & 1.155 & 0.971 & 1.699 \\
$\mathbb F_q^n$ AP-free & $n=5,\ q=5$ & P & 0.804 & 1.000 & 0.644 & 1.000 & 0.773 & 0.995 \\
$\mathbb F_q^n$ AP-free & $n=6,\ q=5$ & P & 0.963 & 1.000 & 0.000$^{\dagger}$ & 1.000 & 0.963 & 1.000 \\
\addlinespace[2pt]
MOLS & $n=10$ & P & 1.000 & 1.000 & 0.500 & 1.000 & 1.000 & 1.000 \\
MOLS & $n=12$ & P & 0.400 & 1.000 & 0.400 & 1.000 & 0.400 & 1.000 \\
MOLS & $n=14$ & P & 0.250 & 1.000 & 0.250 & 1.000 & 0.500 & 1.000 \\
MOLS & $n=15$ & P & 0.500 & 1.000 & 0.500 & 1.000 & 0.500 & 1.000 \\
MOLS & $n=18$ & P & 0.400 & 1.000 & 0.200 & 1.000 & 0.400 & 1.000 \\
MOLS & $n=20$ & P & 0.750 & 1.000 & 0.750 & 0.750 & 0.750 & 0.750 \\
\addlinespace[2pt]
Schur & $k=6$ & P & 0.746 & 1.000 & 0.000$^{\dagger}$ & 1.000 & 0.000$^{\dagger}$ & 1.000 \\
Schur & $k=7$ & P & 0.708 & 1.000 & 0.174 & 1.000 & 0.000$^{\dagger}$ & 1.000 \\
Schur & $k=8$ & P & 0.739 & 1.000 & 0.000$^{\dagger}$ & 1.000 & 0.739 & 1.000 \\
\addlinespace[2pt]
covering & $k=6,\ t=3,\ v=24$ & P & 0.991 & 1.000 & 0.333 & 0.800 & 0.935 & 1.000 \\
covering & $k=6,\ t=3,\ v=30$ & P & 0.818 & 1.000 & 0.000$^{\dagger}$ & 0.784 & 0.000$^{\dagger}$ & 1.000 \\
covering & $k=7,\ t=4,\ v=18$ & P & 0.840 & 1.000 & 0.000$^{\dagger}$ & 1.000 & 0.000$^{\dagger}$ & 1.000 \\
\addlinespace[2pt]
Shannon & $d=24,\ q=7$ & C & 0.515 & 1.009 & 0.515 & 0.991 & 0.713 & 1.009 \\
Shannon & $d=40,\ q=7$ & C & 0.582 & 1.000 & 0.304 & 1.000 & 0.000$^{\dagger}$ & 1.000 \\
Shannon & $d=48,\ q=9$ & C & 1.000 & 1.000 & 1.000 & 1.000 & 1.000 & 1.000 \\
Shannon & $d=64,\ q=9$ & C & 1.000 & 1.000 & 1.000 & 1.000 & 1.000 & 1.000 \\
\addlinespace[2pt]
trifference & $m=1,\ n=64$ & C & 3.000 & 9.000 & 0.000$^{\dagger}$ & 1.000 & 3.000 & 27.000 \\
trifference & $m=1,\ n=96$ & C & 1.000 & 1.000 & 0.000$^{\dagger}$ & 1.000 & 1.000 & 1.000 \\
trifference & $m=1,\ n=144$ & C & 1.000 & 1.000 & 0.000$^{\dagger}$ & 0.000$^{\dagger}$ & 1.000 & 1.000 \\
trifference & $m=1,\ n=192$ & C & 1.000 & 1.000 & 1.000 & 1.000 & 1.000 & 1.000 \\
\end{longtable}